\documentclass{article} 
\usepackage{iclr2026_conference,times}
\iclrfinalcopy

\usepackage{amsmath,amsfonts,bm}

\def\eqref#1{equation~\ref{#1}}

\def\1{\bm{1}}

\DeclareMathAlphabet{\mathsfit}{\encodingdefault}{\sfdefault}{m}{sl}
\SetMathAlphabet{\mathsfit}{bold}{\encodingdefault}{\sfdefault}{bx}{n}

\usepackage[utf8]{inputenc}      
\usepackage[T1]{fontenc}         
\usepackage{CJKutf8}             

\usepackage{amsmath, amssymb, amsfonts, mathrsfs} 
\usepackage{nicefrac}            
\usepackage{microtype}           
\usepackage{xspace}             

\usepackage{graphicx}            
\usepackage{wrapfig}             
\usepackage{subcaption}         
\usepackage{caption}            
\usepackage{tikz}                
\usepackage{asymptote}           
\usepackage{float}              

\usepackage{booktabs}            
\usepackage{multirow}            
\usepackage{makecell}            
\usepackage{array, tabularx, longtable, colortbl, threeparttable} 
\usepackage{arydshln}            

\usepackage{enumitem}            
\usepackage{listings}            
\usepackage{color, xcolor, soul} 

\usepackage{algorithm}           
\usepackage{algorithmicx}        
\usepackage{algpseudocode}       

\usepackage{url}                 
\usepackage{hyperref}            
\usepackage{cleveref}            
\usepackage{listings}
\usepackage{natbib}              

\usepackage{minitoc}             
\usepackage{bbding}              
\usepackage{lscape}              
\usepackage{tablefootnote}

\definecolor{darkred}{RGB}{178,0,0}

\usepackage[colorinlistoftodos, textsize=small]{todonotes}

\newcommand{\glmaw}{80.2\%}
\newcommand{\glmal}{53.6\%}
\newcommand{\qwaw}{72.0\%}
\newcommand{\qwal}{42.5\%}

\title{MobileRL: Online Agentic Reinforcement Learning for  Mobile GUI Agents}

\author{Yifan Xu$^{1*\dagger}$, Xiao Liu$^{1,2*}$, Xinghan Liu$^{1\dagger}$, Jiaqi Fu$^{1\dagger}$, Hanchen Zhang$^{1\dagger}$, Bohao Jing$^{2}$, \\
\textbf{Shudan Zhang$^{1\dagger}$, Yuting Wang$^{2}$,Wenyi Zhao$^{2}$, Yuxiao Dong$^{1}$} \\
\\
\textsuperscript{1} Tsinghua University \quad
\textsuperscript{2} Z.AI \quad
}

\newcommand{\method}{\textsc{MobileRL}\xspace}
\newcommand{\reasoningsft}{Iterative Reasoning Refinement\xspace}
\newcommand{\onlinerl}{\textsc{AdaGRPO}\xspace}
\newcommand{\replaybuffer}{\textsc{AdaPR}\xspace}

\newcommand{\modelglm}{\textsc{AutoGLM}-Mobile-9B\xspace}
\newlength{\ablationboxht}
\newcommand{\hide}[1] 

\begin{document}

\lstset{
  breaklines=true,        
  breakatwhitespace=true, 
  basicstyle=\ttfamily,   
}

\maketitle

\renewcommand{\thefootnote}{\fnsymbol{footnote}}
    \footnotetext[1]{YX and XL contributed equally.}
    \footnotetext[2]{Work partially done while these authors interned at Z.AI.}
\renewcommand{\thefootnote}{\arabic{footnote}}

\vspace{-6mm}

\begin{abstract}

Building general-purpose graphical user interface (GUI) agents has become increasingly promising with the progress in vision language models. 
However, developing effective mobile GUI agents with reinforcement learning (RL) remains challenging due to the heavy-tailed distribution of task difficulty  
and the inefficiency of large-scale environment sampling. 
We present an online agentic reinforcement learning framework \method to enhance GUI agents in mobile environments.  
Its core component is the Difficulty-\textbf{ADA}ptive GRPO (\onlinerl) algorithm. 
In \onlinerl, we design difficulty-adaptive positive replay and failure curriculum filtering to adapt the model to different task difficulties. 
We introduce the shortest-path reward adjustment strategy to reshape rewards concerning the task length in multi-turn agentic tasks. 
Those strategies jointly stabilize RL training, improve sample efficiency, and generate strong performance across diverse mobile apps and tasks. 
We apply \method to two open models (Qwen2.5-VL-7B-Instruct  and GLM-4.1V-9B-Base). 
The resultant \method-9B model achieves state-of-the-art results in terms of success rates on both AndroidWorld~(\glmaw) and AndroidLab~(\glmal). 
The \method framework is open-sourced at \url{https://github.com/THUDM/MobileRL}. 
\end{abstract}

\vspace{-2mm}


\begin{figure*}[h!]
  \centering
  \includegraphics[width=0.8\linewidth]{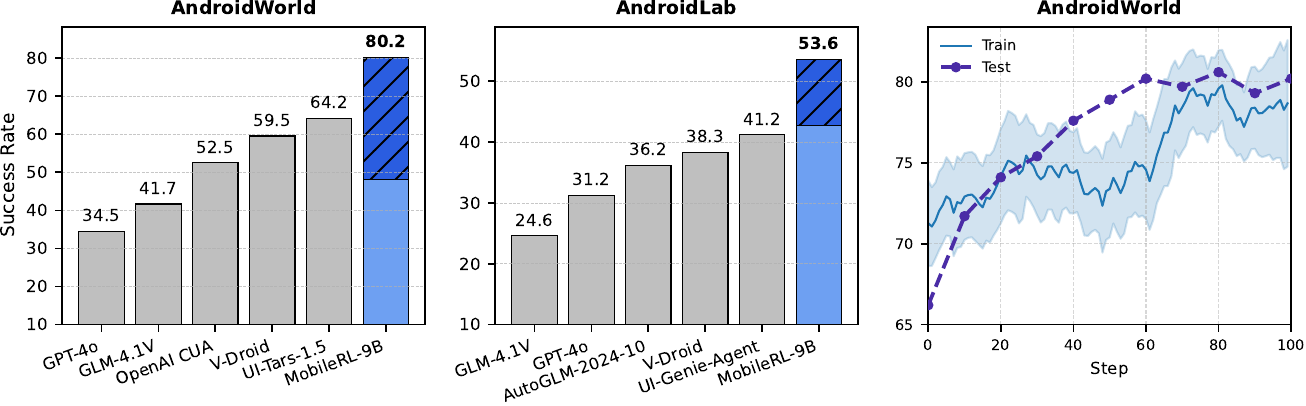}
  \vspace{-2mm}
  \caption{\textit{Left and Center}: Task success rates on AndroidWorld~\citep{rawles2024androidworlddynamicbenchmarkingenvironment} and AndroidLab~\citep{xu2024androidlabtrainingsystematicbenchmarking}; hatched areas indicate gains from \method on top of the SFT model. \textit{Right}: Trajectory-level success rate curves on AndroidWorld train and test sets during RL training.
  }
  \label{fig:android_all}
  \vspace{-3mm}
\end{figure*}

\hide{

\begin{abstract}
Vision language models (VLMs) have recently shown potential as general-purpose agents for graphical user interface (GUI) interaction. 
However, extending them to mobile environments remains challenging due to (i) complex instruction following under limited supervision, (ii) the heavy-tailed distribution of task difficulty, and (iii) the inefficiency of large-scale environment sampling. 
We introduce \textbf{\method}, a unified framework designed to enhance vision language agents in mobile GUI tasks. 
First, \emph{\reasoningsft} converts action-only demonstrations into reasoning--action pairs via an off-the-shelf model and supervised fine-tuning, enabling more effective use of expert data. 
Second, we propose \emph{Difficulty--Adaptive GRPO (\onlinerl)}, an online reinforcement learning algorithm that integrates Difficulty-Adaptive Positive Replay, Failure Curriculum Filtering, and Shortest-Path Reward Adjustment, which jointly stabilize training by adapting to task difficulty and rewarding efficient solutions aligned with user preferences. 
We evaluate our approach by applying \method\ to Qwen2.5-VL-7B-Instruct~\citep{Qwen2.5-VL} and GLM-4.1V-9B-Base~\citep{glmvteam2025glm41vthinkingversatilemultimodalreasoning}. The  \modelglm based on  GLM-4.1V-9B-Base achieves state-of-the-art results with success rates of \textbf{\glmaw}on AndroidWorld~\citep{rawles2024androidworlddynamicbenchmarkingenvironment} and \textbf{\glmal}on AndroidLab~\citep{xu2024androidlabtrainingsystematicbenchmarking}.
The algorithm and framework are adopted in building \textsc{\href{https://autoglm.zhipuai.cn}{AutoGLM}}~\citep{liu2024autoglmautonomousfoundationagents}
\end{abstract}
\vspace{-2mm}

\begin{figure*}[h!]
  \centering
  \includegraphics[width=\linewidth]{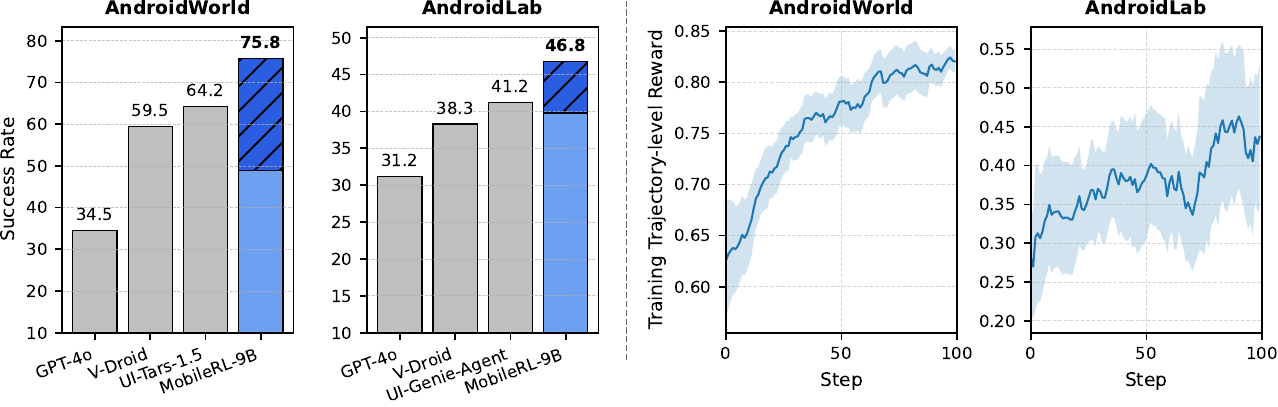}
  \vspace{-2mm}
  \caption{\textbf{Left:} Success rates (SR) on AndroidWorld~\citep{rawles2024androidworlddynamicbenchmarkingenvironment} and AndroidLab~\citep{xu2024androidlabtrainingsystematicbenchmarking}; hatched areas indicate gains from \method. \textbf{Right:} Trajectory-level rewards of \method with 95\% CIs on training sets, showing consistent performance growth. Overall, \method outperforms prior work on both AndroidWorld (w/ rule-based rewards) and AndroidLab (w/ a reward model).}
  \label{fig:android_all}
  \vspace{-3mm}
\end{figure*}

}

\section{Introduction}

GUI agents---powered by vision language models---have enabled zero-shot interaction with web pages and mobile interfaces~\citep{hong2023cogagent,openai-cua,Qwen2.5-VL,liu2024autoglmautonomousfoundationagents}. 
To improve them, significant efforts have focused on supervised fine-tuning or offline imitation learning over static expert demonstrations~\citep{rawles2023aitw,xu2024androidlabtrainingsystematicbenchmarking,bai2024digirltraininginthewilddevicecontrol,lu2025uir1enhancingefficientaction}. 
However, these methods suffer from limited behavior coverage and poor error recovery~\citep{chang2022mitigatingcovariateshiftimitation}. 
Reinforcement learning (RL) with verifiable rewards presents a promising alternative~\citep{deepseekai2025deepseekr1incentivizingreasoningcapability,hou2025t1advancinglanguagemodel}.
Yet, existing datasets with single-step expert labels ~\citep{qin2025uitarspioneeringautomatedgui,luo2025guir1generalistr1style} are insufficient for training or evaluating policies on agentic tasks (i.e., planning and reasoning over multi-step action sequences). 
Although early progress has been made on online learning for GUI agents~\citep{bai2024digirltraininginthewilddevicecontrol,dong2025arpo,dai2025advancingmobileguiagents},  efficiently \textit{scaling agentic RL in interactive mobile simulators} remains largely unexplored.

Specifically, it faces the following technical challenges: 
(i) \textit{Complex instruction following under sparse positive signals}: 
base models usually struggle to reliably produce correct action commands for complex, GUI-specific instructions. 
Due to the heavy cost and latency of mobile emulation, correctly-executed rollouts are rare, resulting in data-inefficient early exploration. 
(ii) \textit{Large and unstable task difficulty spectrum}: 
some tasks can succeed with multiple rollouts, while others are persistently unsolvable for the model. 
Naive sampling wastes computational budget and under-utilizes scarce but informative  trajectories~\citep{xu2024androidlabtrainingsystematicbenchmarking}. 
(iii) \textit{Sampling bottlenecks in large-scale mobile environments}: deploying and managing hundreds of concurrent mobile instances is resource-intensive and hard to reproduce across setups. 
Low sampling throughput further limits both the scale and efficiency of online agentic RL.

To address these challenges, we present an adaptive online agentic RL framework \method for advancing mobile GUI agents. 
\method consists of three components: reasoning-free supervised fine-tuning (SFT), reasoning SFT, and agentic RL. 
The two SFT stages provide a warm-up for RL. 
Specifically, reasoning SFT enhances the handling of long and compositional instructions, reduces costly on-policy trials in mobile simulators, and enables the broad use of open or human-labeled datasets without relying on proprietary models.

To enable effective online agentic RL, we introduce Difficulty–\textbf{ADA}ptive Group Relative Policy Optimization~(\onlinerl). 
Built upon group relative policy optimization (GRPO)~\citep{shao2024deepseekmathpushinglimitsmathematical}, its core idea is to adapt optimization to instance difficulty and explicitly reward solution efficiency. 
\onlinerl designs three key strategies:  
(i) \textit{Difficulty-Adaptive Positive Replay~(AdaPR)} maintains a curated buffer of challenging, high-quality trajectories and balances them with on-policy samples~\citep{mnih2015human, zha2019experience}. 
In sparse-reward mobile environments, difficult successes are rare yet highly informative; replaying them amplifies their learning signal and stabilizes policy updates. 
(ii) \textit{Failure Curriculum Filtering~(FCF)}, as a simplified version of curriculum learning~\citep{matiisen2019teacher,narvekar2020curriculum}, down-weights persistently unsolvable tasks using online difficulty statistics, reallocating computational budget toward feasible instances. 
Given the heavy-tailed difficulty distribution observed in mobile agent  benchmarks~\citep{xu2024androidlabtrainingsystematicbenchmarking,rawles2024androidworlddynamicbenchmarkingenvironment}, pruning hard dead-ends improves sample efficiency while retaining signal from recoverable failures. 
(iii) \textit{Shortest‐Path Reward Adjustment~(SPA)} reshapes the reward function based on completion length, assigning higher returns to shorter solutions. 
Length-sensitive rewards counteract bias toward verbose and better align with user preferences in mobile interaction contexts.

We implement \method in a Verl-based framework~\citep{sheng2024hybridflow}, which supports multi-task, multi-turn agentic RL training. 
Unlike previous Android simulator implementations~\citep{toyama2021androidenv,rawles2024androidworlddynamicbenchmarkingenvironment}---which generally do not support true concurrent execution, our framework sustains high throughput that orchestrates \textit{hundreds of Dockerized Android virtual devices~(AVDs)} across multiple machines. 
This setup enables concurrent interaction with over \textit{1{,}000 environments} while preserving reproducibility. 
Since most open-source benchmarks and simulators are built upon the Android operating system~\citep{toyama2021androidenv,rawles2024androidworlddynamicbenchmarkingenvironment}, this design ensures seamless compatibility and faithful reproduction of environment behaviors.


We train \method on Qwen2.5-VL-7B-Instruct~\citep{Qwen2.5-VL} and GLM-4.1V-9B-Base~\citep{glmvteam2025glm41vthinkingversatilemultimodalreasoning}, producing MobileRL-7B and MobileRL-9B, respectively. 
MobileRL-9B lifts the success rates to \glmaw~ on \textsc{AndroidWorld} and \glmal~ on \textsc{AndroidLab}, significantly outperforming previous state-of-the-art results (64.2\% and 41.2\%, respectively). 
MobileRL-7B outperforms the much larger 72B-parameter models \textsc{UI-Tars-1.5}~\citep{qin2025uitarspioneeringautomatedgui} and \textsc{UI-Genie-Agent}~\citep{xiao2025uigenieselfimprovingapproachiteratively} (e.g., +16\% on \textsc{AndroidWorld}), despite being substantially smaller in scale.
Also, extensive ablation studies demonstrate the effectiveness in the design of \onlinerl. 

In summary, our contributions are as follows:
\begin{itemize}[leftmargin=*,itemsep=0pt,parsep=0.2em,topsep=0.2em,partopsep=0.0em]
\item \textbf{\method Framework \& scalable sampling}: We develop \method with a two-stage warm-up followed by online agentic RL for mobile GUI agents. We further establish a distributed sampling implementation that coordinates \textit{hundreds of Dockerized Android virtual devices~(AVDs)}, enabling reproducible large-scale training on Android benchmarks.

\item \textbf{\onlinerl Algorithm}: We introduce Difficulty–Adaptive Group Relative Policy Optimization (\onlinerl), which extends GRPO with (i) \textit{AdaPR} for replaying challenging successful trajectories, (ii) \textit{FCF} for down-weighting persistently unsolved tasks, and (iii) \textit{SPA} for length-sensitive reward shaping, thereby accounting for instance difficulty and solution efficiency.

\item \textbf{Empirical Results}: Training on Qwen2.5-VL-7B-Instruct and GLM-4.1V-9B-Base yields \textsc{MobileRL-7B} and \textsc{MobileRL-9B}. \textsc{MobileRL-9B} reaches \glmaw~on \textsc{AndroidWorld} and \glmal~on \textsc{AndroidLab}, surpassing previous bests (64.2\% / 41.2\%). 
\end{itemize}

\hide{
In summary, our contributions are as follows:
\begin{itemize}[leftmargin=*,itemsep=0pt,parsep=0.2em,topsep=0.2em,partopsep=0.0em]
    \item \textbf{\method Framework \& scalable sampling}: We propose \method, a framework that combines an iterative reasoning-augmented warm-up with online reinforcement learning for mobile GUI agents. We further establish a distributed sampling implementation that coordinates hundreds of dockerized Android virtual devices via gRPC, lowering early exploration cost, improving instruction adherence, and enabling reproducible large-scale training on Android benchmarks.
    \item \textbf{\onlinerl Algorithm}: We introduce Difficulty–\textbf{ADA}ptive Group Relative Policy Optimization (\onlinerl), which extends GRPO with (i) \textit{AdaPR} for replaying challenging successful trajectories, (ii) \textit{FCF} for down-weighting persistently unsolved tasks, and (iii) \textit{SPA} for length-sensitive reward shaping, thereby accounting for instance difficulty and solution efficiency.
    \item \textbf{Empirical Results}: We demonstrate the effectiveness of our approach on \textsc{AndroidWorld} and \textsc{AndroidLab}. Using Qwen2.5-VL-7B-Instruct as the base model, \method w/ Qwen2.5-VL-7B achieves \qwaw and \qwal, respectively. With GLM-4.1V-9B-Base under the same pipeline, \modelglm attains \textbf{\glmaw}on \textsc{AndroidWorld} and \textbf{\glmal}on \textsc{AndroidLab}.
 We further integrate our pipeline into the \textsc{AutoGLM} product.
\end{itemize}

}


\hide{

\section{Introduction}

Advanced vision–language models (VLMs) have recently shown strong promise as GUI agents~\citep{Qwen2.5-VL,glmvteam2025glm41vthinkingversatilemultimodalreasoning}, enabling zero-shot interaction with web pages and mobile interfaces. Prior work typically relies on supervised fine-tuning or offline imitation learning over static expert demonstrations~\citep{rawles2023aitw,xu2024androidlabtrainingsystematicbenchmarking,bai2024digirltraininginthewilddevicecontrol,lu2025uir1enhancingefficientaction}, which leads to brittle generalization under covariate shift, limited coverage of behaviors, and poor error recovery in novel scenarios~\citep{chang2022mitigatingcovariateshiftimitation}. Inspired by advances in reasoning-model training, reinforcement learning with verifiable rewards has become a compelling alternative~\citep{deepseekai2025deepseekr1incentivizingreasoningcapability}. While single-step expert datasets provide canonical action labels that make single-step RL practical~\citep{lu2025uir1enhancingefficientaction}, they cannot train or evaluate complete action sequences. In GUI agents, nascent online-learning approaches exist~\citep{qiwebrl,dong2025arpo,dai2025advancingmobileguiagents}, but a central open challenge remains: systematically reducing exploration cost in mobile simulators.

Training mobile GUI agents that are robust and efficient in interactive mobile environments remains challenging for three practical reasons. (i) \textit{Complex instruction following under sparse positive signals}: untuned base models struggle to stably produce action commands in the required format for complex, GUI-specific instructions~\citep{xu2024androidlabtrainingsystematicbenchmarking}, and the heavy cost and latency of mobile emulation make such correctly executed rollouts rare, leading to data-inefficient early exploration; the heavy cost and latency of mobil emulation make successful rollouts rare, rendering early exploration data-inefficient. (ii) \textit{Large and unstable difficulty spectrum}: many tasks require multiple rollouts to succeed, while others are persistently unsolvable—naive sampling wastes budget and under-utilizes scarce but informative difficult successes. (iii) \textit{Large-scale mobile environment sampling bottlenecks}: deploying and managing hundreds of concurrent mobile instances is resource-intensive, difficult to reproduce across setups, and often yields low sampling throughput, limiting the scale and efficiency of online reinforcement learning.

Motivated by these challenges, we introduce \textbf{\method}, a unified framework that couples \textit{warm-up} with \textit{large-scale online reinforcement learning} for mobile GUI agents. In the warm-up stage, the base model is fine-tuned on expert demonstrations. Because mobile-use demonstrations typically contain only final action sequences—omitting intermediate reasoning—we propose \reasoningsft, which uses an off-the-shelf instruct model to augment demonstrations with reasoning–action pairs, followed by supervised fine-tuning. This yields two benefits: (i) explicit intermediate reasoning improves handling of long, compositional instructions and reduces costly on-policy trials needed to obtain positive rewards in mobile simulators; and (ii) it enables full utilization of large-scale open-source or human-annotated expert datasets without relying on proprietary models.

Our online reinforcement learning stage introduces \textbf{Difficulty–Adaptive Group Relative Policy Optimization ~(\onlinerl)}, an extension of Group Relative Policy Optimization~(GRPO)~\citep{shao2024deepseekmathpushinglimitsmathematical} that adapts optimization to instance difficulty and explicitly rewards solution efficiency. \onlinerl incorporates three additional mechanisms. (1) \textit{Difficulty-Adaptive Positive Replay~(\replaybuffer)} maintains a curated buffer of challenging, high-quality trajectories and balances them with fresh on-policy samples. In sparse-reward mobile environments, difficult successes are rare yet highly informative; replaying them amplifies their learning signal and stabilizes policy updates. (2) \textit{Failure Curriculum Filtering~(FCF)} down-weights persistently unsolvable tasks using online difficulty statistics, reallocating computational budget toward challenging but feasible instances. Given the heavy-tailed difficulty distribution observed in mobile phone use benchmarks~\citep{xu2024androidlabtrainingsystematicbenchmarking,rawles2024androidworlddynamicbenchmarkingenvironment}, pruning hard dead-ends improves sample efficiency while retaining signal from recoverable failures. (3) \textit{Shortest‐Path Reward Adjustment~(SPA)} reshapes the reward function based on completion length, granting higher returns to shorter solutions. Length-sensitive rewards counteract the bias toward verbose and better align with user preferences in mobile interaction contexts. 

\begin{figure*}[t]
    \centering
    \begin{subfigure}[t]{0.30\textwidth}
        \centering
        \includegraphics[width=\linewidth]{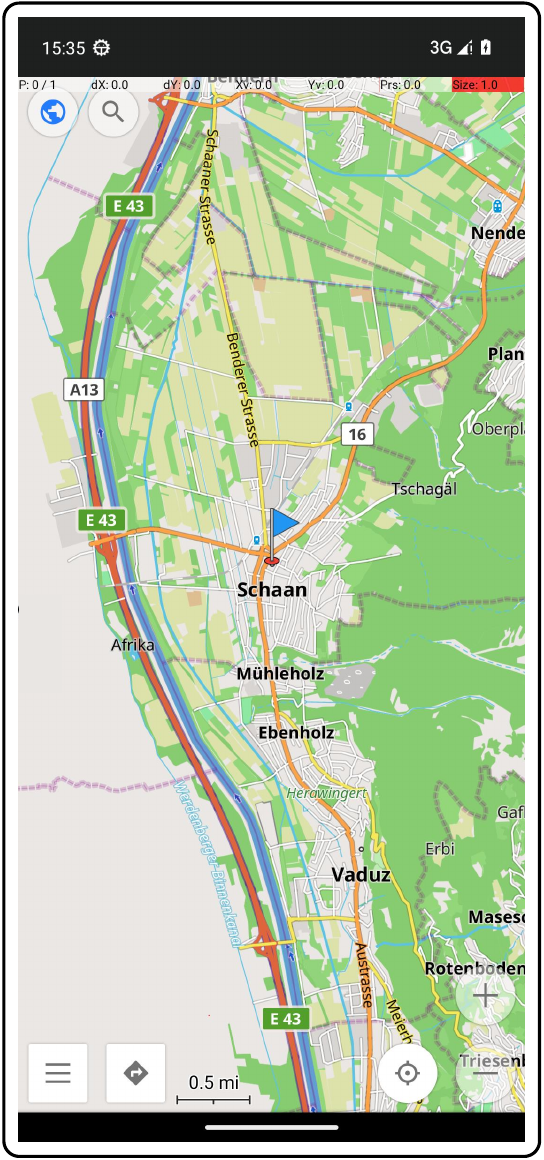}
        \caption{Add a location marker for 47.16, 9.51 in the OsmAnd maps app.}
        \label{fig:sub1}
    \end{subfigure}\hfill
    \begin{subfigure}[t]{0.30\textwidth}
        \centering
        \includegraphics[width=\linewidth]{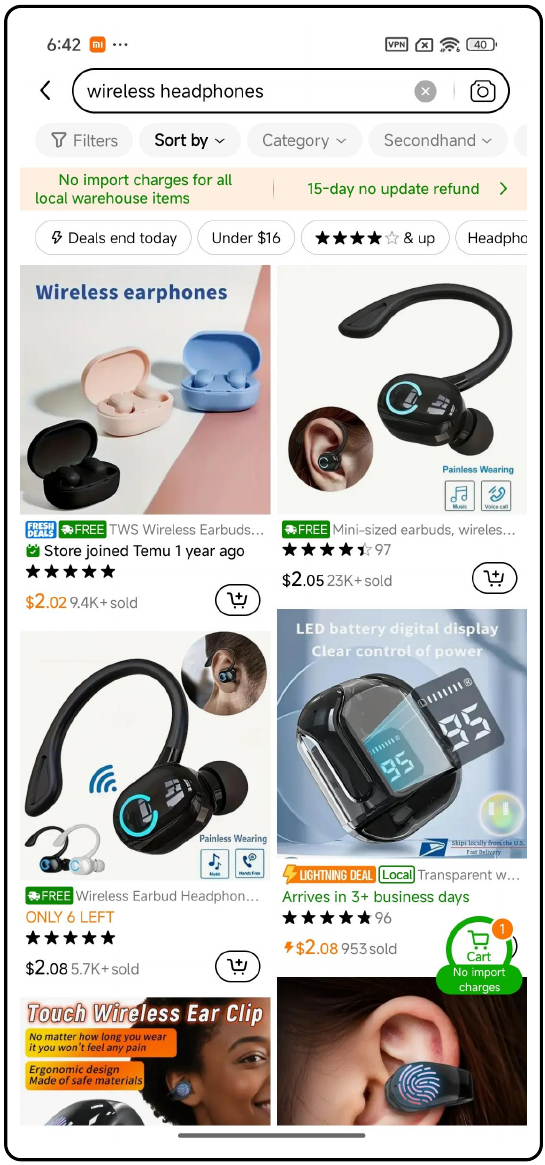}
        \caption{Search for wireless headphones in temu, and sort by price low to high.}
        \label{fig:sub3}
    \end{subfigure}\hfill
    \begin{subfigure}[t]{0.30\textwidth}
        \centering
        \includegraphics[width=\linewidth]{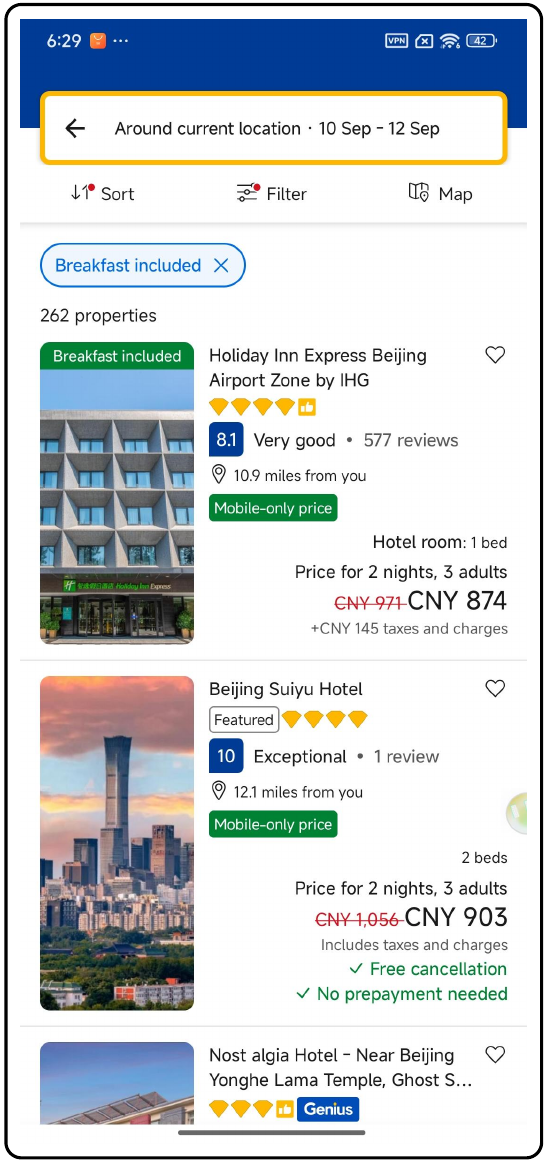}
        \caption{Search for hotels on Booking, check-in date is 09-10, check-out date is 09-12, sort by prices.}
        \label{fig:sub4}
    \end{subfigure}
    \caption{Example mobile tasks finished by our agent. Our agent can automatically perform tasks according to human instructions in academic benchmarks and real-world applications.}
    \label{fig:all}
\end{figure*}

We employ a distributed sampling framework to sustain high throughput in which a gRPC controller orchestrates hundreds of Dockerized Android virtual devices~(AVDs) across multiple machines. This setup enables concurrent interaction with over $1000$ environments while preserving reproducibility. Since most open-source benchmarks and simulators are built upon the Android operating system~\citep{toyama2021androidenv,rawles2024androidworlddynamicbenchmarkingenvironment}, this design ensures seamless compatibility and faithful reproduction of environment behaviors.

We evaluate the effectiveness of the proposed \method framework on two model backbones: Qwen2.5-VL-7B-Instruct~\citep{Qwen2.5-VL} and GLM-4.1V-9B-Base~\citep{glmvteam2025glm41vthinkingversatilemultimodalreasoning}. Experiments are conducted on two interactive mobile use benchmarks: \textsc{AndroidWorld}~\citep{rawles2024androidworlddynamicbenchmarkingenvironment} and \textsc{AndroidLab}~\citep{xu2024androidlabtrainingsystematicbenchmarking}. For a direct comparison under the same Qwen2.5-VL backbone~(7B variant, which is substantially smaller than the $72$B models used by \textsc{UI-Tars-1.5}~\citep{qin2025uitarspioneeringautomatedgui} and \textsc{UI-Genie-Agent}~\cite{xiao2025uigenieselfimprovingapproachiteratively}), \method~w/ Qwen2.5-VL-7B demonstrates significantly better performance. Specifically, \textsc{UI-Tars-1.5} achieves $64.2\%$ on \textsc{AndroidWorld} and \textsc{UI-Genie-Agent} attains $41.2\%$ on \textsc{AndroidLab}, whereas \method~w/ Qwen2.5-VL-7B reaches \qwaw~on \textsc{AndroidWorld} and \qwal~on \textsc{AndroidLab}, resulting in absolute gains of $+7.8$ and $+1.3$ percentage points, respectively. Furthermore, when adopting the GLM-4.1V-9B backbone, \method~w/ GLM-4.1V-9B-Base achieves even stronger results, attaining \glmaw~on \textsc{AndroidWorld} and \glmal~on \textsc{AndroidLab}. These scores represent new state-of-the-art results, with absolute improvements of $+11.6$ and $+5.6$ percentage points over the strongest prior baselines.  We refer to this variant as \modelglm.

In summary, our contributions are as follows:
\begin{itemize}[leftmargin=*,itemsep=0pt,parsep=0.2em,topsep=0.2em,partopsep=0.0em]
    \item \textbf{\method Framework \& scalable sampling}: We propose \method, a framework that combines an iterative reasoning-augmented warm-up with online reinforcement learning for mobile GUI agents. We further establish a distributed sampling implementation that coordinates hundreds of dockerized Android virtual devices via gRPC, lowering early exploration cost, improving instruction adherence, and enabling reproducible large-scale training on Android benchmarks.
    \item \textbf{\onlinerl Algorithm}: We introduce Difficulty–Adaptive Group Relative Policy Optimization (\onlinerl), which extends GRPO with (i) \textit{AdaPR} for replaying challenging successful trajectories, (ii) \textit{FCF} for down-weighting persistently unsolved tasks, and (iii) \textit{SPA} for length-sensitive reward shaping, thereby accounting for instance difficulty and solution efficiency.
    \item \textbf{Empirical Results}: We demonstrate the effectiveness of our approach on \textsc{AndroidWorld} and \textsc{AndroidLab}. Using Qwen2.5-VL-7B-Instruct as the base model, \method w/ Qwen2.5-VL-7B achieves \qwaw and \qwal, respectively. With GLM-4.1V-9B-Base under the same pipeline, \modelglm attains \textbf{\glmaw}on \textsc{AndroidWorld} and \textbf{\glmal}on \textsc{AndroidLab}.
 We further integrate our pipeline into the \textsc{AutoGLM} product.
\end{itemize}

}
\section{\textsc{\method}}

\begin{figure*}[t]
    \centering
    \includegraphics[width=\linewidth]{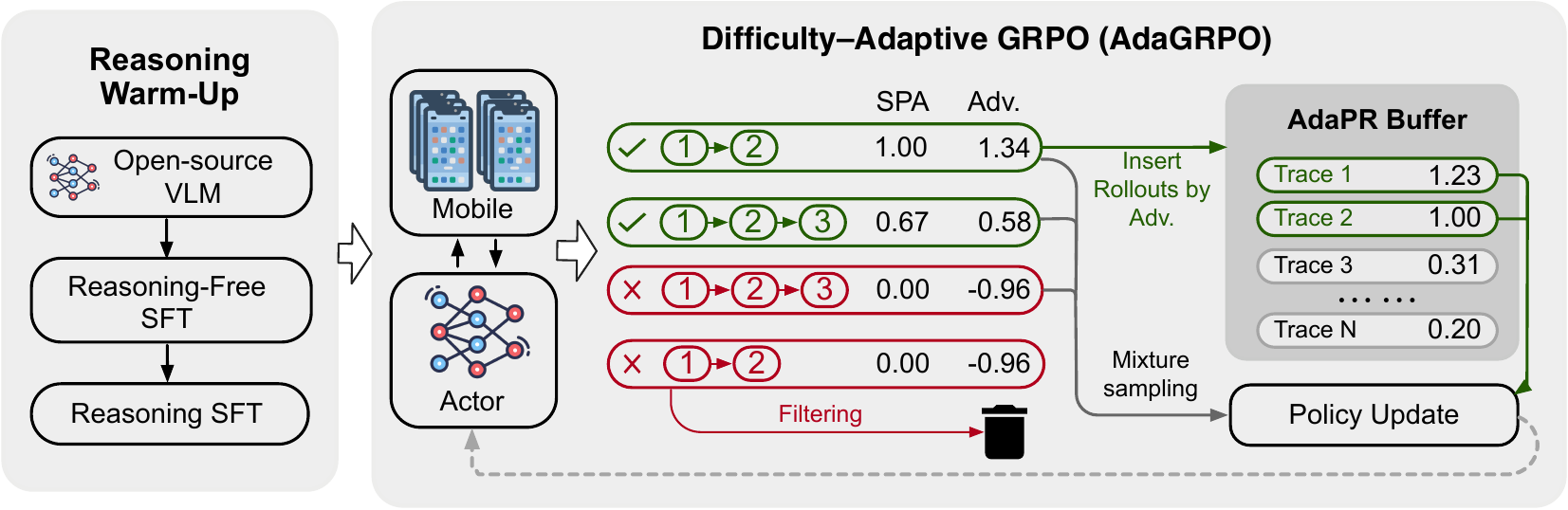}
    \caption{Overview of \method. 
    It consists of 1) reasoning warm-up with both reasoning-free SFT and reasoning SFT and 2) online agentic RL with \onlinerl. 
    In \onlinerl, the warmed-up policy interacts with mobile environments to generate rollouts, which are scored by shortest-path reward adjustment~(SPA). 
    High-quality positive trajectories are stored in the AdaPR buffer, while low-performing rollouts are pruned via failure curriculum filtering.}
    \label{pic:main}

\end{figure*}

We study mobile GUI agents and introduce the \method framework, as shown in Figure~\ref{pic:main}, which aims to address three key challenges in interactive mobile environments: 
(i) following complex instructions under sparse and delayed rewards; 
(ii) handling a heavy–tailed and unstable task difficulty distribution; 
and (iii) overcoming large–scale sampling bottlenecks in mobile emulators. 

Given a natural–language instruction (e.g., ``open the calendar and add an event for tomorrow at 3\,pm''), the agent autonomously performs closed–loop interactions with the mobile device. 
First, it perceives the current screen, grounds UI elements, and executes a sequence of actions without human intervention.
The feedback is sparse and it can be observed only upon successful task completion, at which point the interaction terminates or a predefined horizon is reached.
The goal is to learn a policy that generates strong performance across applications and tasks, minimizes unnecessary interactions, and maximizes task success.

\hide{
We study mobile \emph{UI agents} and introduce the \method framework, which directly targets three practical challenges in interactive mobile environments: (i) following complex instructions under sparse, delayed rewards; (ii) coping with a heavy–tailed, unstable difficulty distribution across tasks; and (iii) overcoming large–scale sampling bottlenecks in mobile emulators.
Given a natural–language instruction (e.g., ``open the calendar and create an event for tomorrow at 3\,pm''), the agent autonomously performs closed–loop interaction with the device: it perceives the current screen, grounds UI elements, and executes a sequence of actions (e.g., \texttt{Tap}, \texttt{Swipe}, \texttt{Type}) without further human intervention.
Feedback is sparse and observed only upon successful task completion, at which point the interaction terminates (or when a predefined horizon is reached).
The goal is to learn a policy that generalizes across diverse apps and screens, maximizes task success, and minimizes unnecessary interactions.
}

\paragraph{Problem Formulation.}
We model the mobile GUI agent as a finite-horizon Markov Decision Process (MDP)~\citep{littman2009tutorial}
\(\mathcal{M}=(\mathcal{S},\mathcal{A},P,r,H,\mu_0)\).
The state space \(\mathcal{S}\) contains all possible GUI states; a concrete state at time \(t\) is \(s_t\in\mathcal{S}\), which comprises a screenshot of the device screen together with the structured UI hierarchy parsed from XML metadata~(Cf. Appendix~\ref{appendix:xml-processing}).
The action space \(\mathcal{A}\) is a finite set of atomic GUI operations; an action at time \(t\) is \(a_t\in\mathcal{A}\), including primitives such as \texttt{Tap}, \texttt{Swipe}, \texttt{Type},  \texttt{Long Press}, \texttt{Launch}, \texttt{Home}, \texttt{Back}, \texttt{Wait}, and a terminating action \texttt{Finish}.
$P(s_{t+1} \mid s_t, a_t)$ represents the stochastic transition mechanism of the Android OS and installed applications, capturing the uncertainty over possible next states.
We consider a finite horizon \(H\in\mathbb{N}\).
The initial condition is drawn from a joint distribution \(\mu_0\) over instruction--state pairs, i.e., \((s_0,c)\sim\mu_0\).
Here, \(c\) is the natural-language task instruction provided \emph{once} at the beginning of the episode (\(t{=}0\)).
At each timestep \(t=0,1,\dots\), the agent observes only the current state \(s_t\) and samples an action according to a policy
\(\pi_\theta\), i.e.,
\(a_t \sim \pi_\theta(\cdot \mid s_t,c)\).
The environment then transitions to \(s_{t+1}\sim P(\cdot\mid s_t,a_t)\).
An episode yields a trajectory $\tau=\big((s_0,a_0),(s_1,a_1),\ldots,(s_T,a_T)\big)$ and terminates either when the agent intentionally selects \texttt{Finish} in a success state or when the horizon is reached, with $T\le H$. The reward is assigned only after task completion, such that $R(\tau)=r(s_T,a_T)$ with $r(s_T,a_T)\in\{0,1\}$ indicating success (1) or failure (0). Consequently, learning maximizes the success probability:~$\theta^\star = \arg\max_{\theta}\;\mathbb{E}_{(s_0,c)\sim\mu_0,\;\tau\sim\pi_\theta}\!\bigl[R(\tau)\bigr]$.

\subsection{The \method Framework}

To build a powerful mobile use agent, we present the \method framework. 
It comprises three components: reasoning-free supervised fine-tuning (SFT) on expert demonstration data, an iterative warm-up stage by reasoning SFT, and agentic RL with a difficulty–adaptive policy optimization strategy we developed in this work. 

\paragraph{Reasoning-Free SFT.} 
In agentic RL training, sampling in virtual-device environments is usually inefficient; thus, starting online RL directly from a base model was found to be excessively time-consuming. Therefore, we perform SFT with expert demonstration data obtained by following the data collection protocol of~\citep{xu2024androidlabtrainingsystematicbenchmarking}, supplemented with the training split of the publicly available AndroidControl dataset~\citep{li2024effects}. Note that this data is reasoning-free.

\paragraph{Reasoning SFT.}
To further construct a stronger reasoning policy initializer, we perform reasoning SFT via an iterative reasoning refinement strategy over the expert dataset. 
Manually collected expert demonstrations dataset for mobile use often contains only the final action sequence, omitting intermediate reasoning.
Training solely on such ``black-box'' trajectories yields opaque policies.
We leverage an off-the-shelf instruction model to bootstrap a reasoning-augmented training set from raw demonstrations, yielding a structured and transparent policy initialization. Concretely, we iteratively build the reasoning instruction–tuning pairs in three stages:
\begin{itemize}[leftmargin=*,itemsep=0pt,parsep=0.2em,topsep=0.2em,partopsep=0.0em]
    \item Bootstrap sampling: For each task \(x\) with expert answer \(a^*\), the Instruct model \(M\) generates diverse candidate reasoning–action pairs \((c_k,a_k)\). Whenever \(a_k=a^*\), we retain \((x,c_k,a^*)\) in \(\mathcal{D}_R\).
    \item Supervised fine-tuning: Train an initial reasoning policy \(\pi^R_0\) on \(\mathcal{D}_R\).
    \item Iterative refinement: At iteration \(t\), \(\pi^R_t\) proposes candidates; those matching \(a^*\) are scored by correctness. The best explanation \(c^*\) is added to \(\mathcal{D}_{\mathrm{new}}\), and \(\pi^R_{t+1}\) is obtained by fine-tuning. 
\end{itemize}

The resulting reasoning-oriented fine-tuning corpus is trained for two epochs to produce the reasoning warm-up model used for agentic RL training. 

\paragraph{Agentic RL.}
During agentic RL (multi-turn) training, we face the challenges of immediate reward assignment and sampling efficiency, which are discussed and addressed by building upon the group relative policy optimization (GRPO)~\citep{shao2024deepseekmathpushinglimitsmathematical}. 

Briefly, GRPO advances proximal policy optimization~\citep{schulman2017proximalpolicyoptimizationalgorithms} by replacing the learned value baseline with an on-the-fly, group-relative baseline computed from a set of trajectories for the same task. 
Given a shared initial condition $(s_0,c)\sim \mu_0$, we sample a group of $G$ trajectories
$\mathcal{G}=\{\tau_1,\ldots,\tau_G\}$ by rolling out $\pi_{\theta_{\mathrm{sample}}}$.
Let $T_i$ denote the number of steps for trajectory $\tau_i$, 
$s_{i,t},a_{i,t}$ be the state and action at step $t$, and the trajectory-level reward for $\tau_i$ be $R(\tau_i)\in\{0,1\}$.
We define the group-relative trajectory-level advantage for any step $(s_{i,t},a_{i,t})$ on trajectory $\tau_i$ as
$
\hat A_{i,t}\;\equiv\;
\frac{R(\tau_i)-\mu_R}{\sigma_R} \,
\label{eq:advantage_grpo}
$
, where $\mu_R$ and $\sigma_R$ are the mean and standard deviation of $\{R(\tau_j)\}_{j=1}^G$.
The GRPO loss can be written in empirical form as
\begin{equation}
\begin{aligned}
\mathcal{L}^{\mathrm{GRPO}}(\theta)=
-\,\frac{1}{G}\sum_{i=1}^G \frac{1}{T_i}\sum_{t=1}^{T_i}
\min\!\Bigl(\rho_{i,t}(\theta)\,\hat A_{i,t},\;
\mathrm{clip}\!\bigl(\rho_{i,t}(\theta),1-\epsilon,1+\epsilon\bigr)\,\hat A_{i,t}\Bigr)
\;
\end{aligned}
\label{eq:loss_grpo_empirical}
\end{equation}
where  
$
\rho_{i,t}(\theta)
=\frac{\pi_{\theta}(a_{i,t}\mid s_{i,t})}
       {\pi_{\theta_{\mathrm{sample}}}(a_{i,t}\mid s_{i,t})}
$ is the token-wise importance sampling (IS) ratio. We add the KL loss $\beta\,D_{\mathrm{KL}}\!\bigl(\pi_{\theta}\,\|\,\pi_{\theta_{\mathrm{ref}}}\bigr)$ to prevent the model from deviating too much from the prior distribution.

\subsection{Difficulty–Adaptive GRPO}
\label{subsec: dgrpo}

We develop Difficulty–Adaptive GRPO (\onlinerl) with three strategies---shortest-path reward adjustment (SPA), difficulty–adaptive positive replay (AdaPR), and failure curriculum filtering (FCF)---to address the challenges faced in mobile agentic RL. 

First, in multi‐turn mobile agentic tasks—where immediate rewards are absent within a single round, unlike single‐turn settings~\citep{lu2025uir1enhancingefficientaction,glmvteam2025glm41vthinkingversatilemultimodalreasoning}—the reward allocation strategy must be redesigned. Beyond assigning a uniform terminal reward, we introduce \emph{Shortest-Path Reward Adjustment (SPA)}, which reshapes rewards with respect to task length. The goal of the adjustment is to provide more informative learning signals, guiding the model toward accurate and efficient completion paths and facilitating the computation of trajectory-level advantages.

Second, a further challenge arises from the uniform sampling strategy employed in standard GRPO. 
In mobile use scenarios---where each sample carries a significant computational cost—this approach results in poor sample efficiency, particularly due to the repeated inclusion of inherently unsolvable tasks. 
To mitigate this, we adapt data collection and training based on instance difficulty through two mechanisms: {Difficulty–Adaptive Positive Replay (AdaPR)} and {Failure Curriculum Filtering (FCF)}. 
At the same time, we restrict redundant successful trajectories to avoid unnecessary updates and promote training efficiency.

\subsubsection{Shortest‐Path Reward Adjustment (SPA)}

In mobile tasks, the environment returns a binary terminal reward 
\(r \in \{0,1\}\)~\citep{xu2024androidlabtrainingsystematicbenchmarking,rawles2024androidworlddynamicbenchmarkingenvironment} indicating task success.  
Previous approaches typically broadcast this reward to every timestep, i.e.,  
\(R(s_t,a_t)=r,\; t=0,\dots,T\),  
so that the per-step signal remains aligned with the sparse objective.  
However, assigning identical rewards to all successful rollouts biases training
toward \emph{longer} trajectories, since they contribute more gradient terms.
To counteract this, we introduce SPA,  
which re-scales the reward for each trajectory \(\tau_i\) as
\begin{equation}
R^{\mathrm{SPA}}(s_t,a_t) = r(\tau_i)\left(1 - \alpha \,\frac{T_i - T_{\min}}{T_i}\right),
\qquad T_{\min}=\min_{\tau_j \in \mathcal{T}_{\mathrm{succ}}}|\tau_j|,\;\alpha\in(0,1].
\end{equation}

where \(T_i = |\tau_i|\) is the length of trajectory \(\tau_i\), and 
\(\mathcal{T}_{\mathrm{succ}}=\{\tau_j \mid r(\tau_j)=1\}\) denotes the set of 
\emph{successful} trajectories for the current problem instance.
Here \(T_{\min}\) is the length of the shortest successful trajectory in \(\mathcal{T}_{\mathrm{succ}}\), 
and \(\alpha\!\in(0,1]\) controls the penalty strength.
In this formulation, shorter sequences are not automatically considered better; 
unsuccessful early terminations still receive a reward of \(0\). 
This adjustment encourages the policy to prefer shorter, 
successful paths without sacrificing the success rate.



\subsubsection{Difficulty–Adaptive Positive Replay (AdaPR)}
In sparse-reward mobile environments, successful yet challenging rollouts are rare but highly informative; effectively leveraging them enhances the learning signal and stabilizes policy updates. 
Therefore, inspired by the paradigm of experience replay~\citep{mnih2015human, zha2019experience}, we introduce difficulty-adaptive positive replay (AdaPR) strategically to retain and reuse challenging, high-value trajectories while blending them with fresh on-policy samples.
We introduce key components of AdaPR: buffer construction for high-quality trajectory selection and mixture sampling to balance replay and exploration.

\paragraph{Buffer Construction.}
At iteration $t$, the rollout set is
$\mathcal{T}_t=\{\tau^{(1)},\dots,\tau^{(N)}\}$, collected under
the current policy $\pi_{\theta_t}$.
We compute the trajectory-level advantage and insert the top $\kappa$ trajectories into the replay buffer $\mathcal{B}$.

\paragraph{Mixture Sampling.}
Each policy update is performed on a mini‑batch of $M$ trajectories obtained
from the mixture distribution
$
    \label{eq:mixture}
    q(\tau)\;=\;\gamma\,p_{\mathcal{B}}(\tau)
              \;+\;(1-\gamma)\,p_{\mathrm{on}}(\tau)
$
where $p_{\mathrm{on}}$ is the on‑policy distribution
$\pi_{\theta_t}$ and $p_{\mathcal{B}}$ is the empirical distribution over
$\mathcal{B}$.
To keep the replay contribution under control, at most
$\gamma M$ trajectories with the highest current
advantage are drawn from $\mathcal{B}$, preserving on‑policy diversity.

\subsubsection{Failure Curriculum Filtering (FCF)}

To avoid repeatedly sampling tasks that yield zero reward—which wastes computation and hinders the collection of positive advantage data, we propose failure curriculum filtering. 
In FCF, any task producing all-zero rewards for two consecutive epochs enters a three-epoch cooldown, during which its sampling probability is reduced according to $w_{\text{task}} = \exp(-f)$, where $f$ is the number of consecutive failure epochs; after this cooldown period, the task is permanently removed.
This method can be regarded as a variant of curriculum sampling~\citep{matiisen2019teacher,narvekar2020curriculum}. To avoid excessive hyperparameter tuning, we simplify it to target only the exclusion of the most difficult subset of data.
Tasks with consistently high failure counts are permanently removed from the sampling pool. 
For stability, failure histories from previous training are retained.

\paragraph{Summary.}
In summary, \method consists of reasoning-free SFT, reasoning SFT, and difficulty-adaptive RL for training mobile GUI agents. 
Reasoning-free SFT helps build a strong action foundation from expert demonstrations, while reasoning SFT adds intermediate reasoning to improve instruction following and policy transparency. 
On top of this initialization, agentic RL with \onlinerl addresses the challenges of sparse terminal rewards, heavy-tailed task difficulty, and expensive sampling. 
Specifically, SPA reshapes terminal rewards for denser feedback, AdaPR strategically reuses challenging successful trajectories, and FCF filters out persistently-unsolvable tasks.

\hide{

\section{\textsc{\method}}

\begin{figure*}[t]
    \centering
    \includegraphics[width=\linewidth]{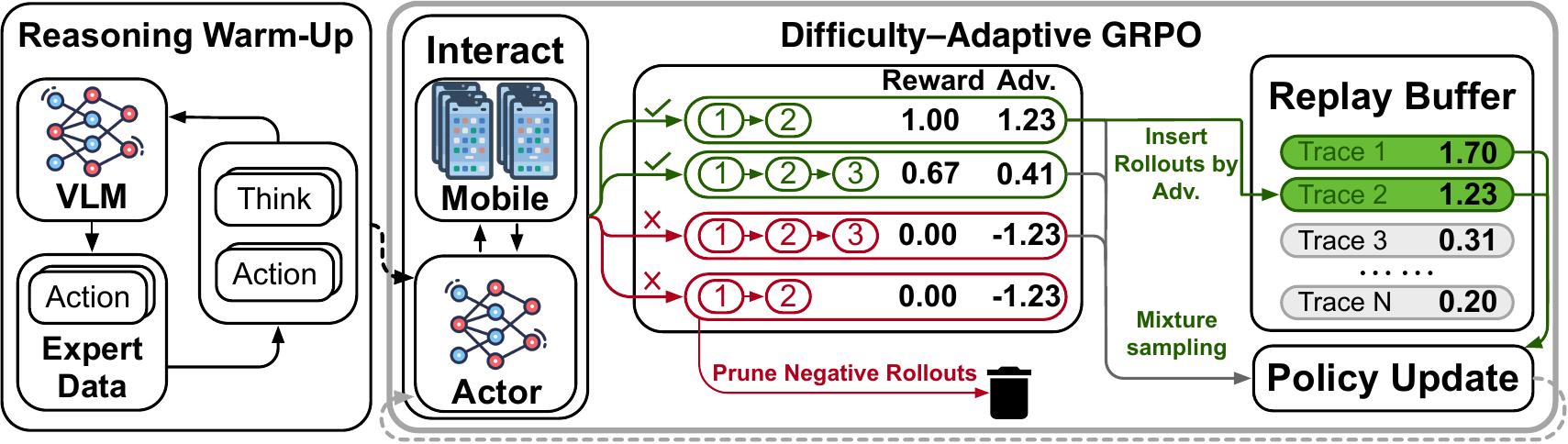}
    \caption{Overview of \method. (1) \emph{\reasoningsft:} the model performs multi-round, self-bootstrapped rollouts on expert data to construct a reasoning-augmented initializer. (2) \emph{Online training with \onlinerl:} the warmed-up policy is refined via interaction; top positive trajectories enter a replay buffer while part of the negatives are pruned. Policy updates samples from a mixture of on-policy data and the buffer.}

\end{figure*}


The \method framework comprises three components: supervised fine-tuning on expert demonstration data, an iterative warm-up stage \emph{\reasoningsft}, and \emph{Difficulty–Adaptive GRPO} (\onlinerl). Because sampling in virtual-device environments is inefficient, starting online reinforcement learning directly from a base model was found to be excessively time-consuming in preliminary experiments. We begin by following the data collection protocol of~\cite{xu2024androidlabtrainingsystematicbenchmarking} to obtain expert demonstrations, which are then used for supervised fine-tuning.
We then construct a stronger reasoning initializer via \reasoningsft over the expert dataset, followed by the application of \onlinerl for efficient online refinement.

For the observation space, we adopt a dual presentation: the current screenshot and a compressed Extensible Markup Language~(XML). The preprocessing steps for XML simplification are detailed in Appendix~\ref{appendix:xml-processing}.
In most cases, the agent can use coordinates from XML to specify click positions, bypassing brittle pixel-level grounding; when graphical cues or incomplete XML are involved, the screenshot provides the necessary visual details.

\paragraph{Problem Setting}
We model device control and GUI navigation under natural-language instructions as a finite-horizon Markov Decision Process (MDP)~\citep{littman2009tutorial}, \(\mathcal{M}=(\mathcal{S},\mathcal{A},P,R,H)\).
Here, \(\mathcal{S}\) is the device state; \(\mathcal{A}\) is a discrete action space covering \texttt{Tap}, \texttt{Swipe}, \texttt{Type}, \texttt{Long~Press}, and \texttt{Finish}; \(P(s_{t+1}\!\mid\!s_t,a_t)\) denotes environment dynamics induced by Android operating system and applications; \(R(s_t,a_t)\) is a sparse reward that yields \(1\) iff the task is completed at step \(T\) and \(0\) otherwise; and \(H\) is the interaction horizon.
Given an initial state–instruction pair \((s_0,c)\sim\mu_0\), an episode forms a trajectory \(\tau=(s_0,a_0,\dots,s_T)\) and terminates when the task is fulfilled or the horizon \(H\) is reached.

\paragraph{GRPO}
\label{sec:grpo}

Group Relative Policy Optimization (GRPO)~\citep{shao2024deepseekmathpushinglimitsmathematical}
extends Proximal Policy Optimization by replacing the learned value baseline
with an on-the-fly, group-relative baseline computed from a set of trajectories
for the same task. Given a state $s$, we sample a group of $G$ actions
$a_1,\ldots,a_G \sim \pi_{\theta_{\mathrm{old}}}(\cdot \mid s)$ and define the
group-relative advantage using the reward function $R$:
\[
A^{\mathrm{GRPO}}(s,a_i)
=\frac{R(s,a_i)-\mathrm{avg}_{j=1}^G\,R(s,a_j)}
       {\mathrm{std}_{j=1}^G\,R(s,a_j)}\, .
\]
Let $T_i$ denote the number of steps for sample $i$ and
$s_{i,t},a_{i,t}$ be the state and action at step $t$.
Define the per-step importance ratio and the per-step advantage as
\[
\rho_{i,t}(\theta)
=\frac{\pi_{\theta}(a_{i,t}\mid s_{i,t})}
       {\pi_{\theta_{\mathrm{old}}}(a_{i,t}\mid s_{i,t})},
\qquad
\hat A_{i,t}\;\equiv\;A^{\mathrm{GRPO}}(s_{i,t},a_{i,t}).
\]
The policy maximizes the clipped surrogate
\begin{align*}
J^{\mathrm{GRPO}}(\theta)=
\mathbb{E}_{\substack{s\sim d_{\pi_{\theta_{\mathrm{old}}}},\\ a_{1:G}\sim \pi_{\theta_{\mathrm{old}}}(\cdot\mid s)}}\Biggl[
\frac{1}{G}\sum_{i=1}^G \frac{1}{T_i}\sum_{t=1}^{T_i}
\min\!\Bigl(\rho_{i,t}(\theta)\,\hat A_{i,t},\;
\mathrm{clip}\!\bigl(\rho_{i,t}(\theta),1-\epsilon,1+\epsilon\bigr)\,\hat A_{i,t}\Bigr)
\;\\-\;
\beta\,D_{\mathrm{KL}}\!\bigl(\pi_{\theta}\,\|\,\pi_{\theta_{\mathrm{ref}}}\bigr)
\Biggr],
\end{align*}
where $D_{\mathrm{KL}}(p\|q)=\sum_{a} p(a)\log\frac{p(a)}{q(a)}$ and
$d_{\pi_{\theta_{\mathrm{old}}}}$ denotes the state visitation distribution.

\subsection{\reasoningsft~(IRR)}
\label{sec:rbir}

Manually collected expert demonstrations dataset for mobile use often contains only the final action sequence, omitting intermediate reasoning.
Training solely on such ``black-box'' trajectories yields opaque policies, while many unlabeled tasks remain unused.
We leverage an off-the-shelf Instruct model to activate expert data and bootstrap a reasoning-augmented training set from raw demonstrations, yielding a structured and transparent policy initialization.

Concretely, we iteratively build reasoning instruction–tuning pairs in three stages:
\begin{itemize}[leftmargin=*,itemsep=0pt,parsep=0.2em,topsep=0.2em,partopsep=0.0em]
    \item Stage~0 (Bootstrap sampling): For each task \(x\) with expert answer \(a^*\), the Instruct model \(M\) generates diverse candidate reasoning–action pairs \((c_k,a_k)\) (via, e.g., temperature/nucleus sampling). Whenever \(a_k=a^*\), we retain \((x,c_k,a^*)\) in \(\mathcal{D}_R\).
    \item Stage~1 (Supervised fine-tuning): Train an initial reasoning policy \(\pi^R_0\) on \(\mathcal{D}_R\).
    \item Stage~2 (Iterative refinement): At iteration \(t\), \(\pi^R_t\) proposes candidates; those matching \(a^*\) are scored by conciseness and \(\log P(c)\). The best explanation \(c^*\) is added to \(\mathcal{D}_{\mathrm{new}}\), and \(\pi^R_{t+1}\) is obtained by fine-tuning. We stop when the match rate saturates.
\end{itemize}

\subsection{Difficulty–Adaptive GRPO~(\onlinerl)}

Building on GRPO, we propose \textbf{D}ifficulty–Adaptive GRPO (\onlinerl) tailored to our tasks.
In multi-turn mobile use tasks, where immediate rewards are not available within a single round as in single-turn tasks~\citep{lu2025uir1enhancingefficientaction,glmvteam2025glm41vthinkingversatilemultimodalreasoning}, the reward allocation strategy must be carefully redesigned. Beyond simply assigning a uniform final reward at the trajectory level, we introduce the \emph{Shortest-Path Reward Adjustment}, which reshapes rewards concerning task length. This adjustment provides more informative learning signals, guiding the model toward accurate and efficient completion paths and facilitating the computation of trajectory-level advantages.
This also facilitates the computation of trajectory-level advantages tailored to sparse-reward settings.

A further challenge arises from the uniform sampling strategy employed in standard GRPO. In mobile use scenarios—where each sample carries a significant computational cost—this approach results in poor sample efficiency, particularly due to the repeated inclusion of inherently unsolvable tasks. To mitigate this, we adapt data collection and training based on instance difficulty through two mechanisms: \emph{Difficulty–Adaptive Positive Replay} and \emph{Failure Curriculum Filtering}. At the same time, we restrict redundant successful trajectories to avoid unnecessary updates and promote training efficiency.

\subsubsection{Shortest‐Path Reward Adjustment (SPA).}

In mobile tasks, the environment returns a binary terminal reward 
\(r \in \{0,1\}\)~\citep{xu2024androidlabtrainingsystematicbenchmarking,rawles2024androidworlddynamicbenchmarkingenvironment} indicating task success.  
Previous approaches typically broadcast this reward to every timestep, i.e.,  
\(R(s_t,a_t)=r,\; t=0,\dots,T\),  
so that the per-step signal remains aligned with the sparse objective.  
However, assigning identical rewards to all successful rollouts biases training
toward \emph{longer} trajectories, since they contribute more gradient terms.
To counteract this, we introduce SPA,  
which re-scales the reward for each trajectory \(\tau_i\) as
\begin{equation}
    R^{\mathrm{SPA}}(s_t,a_t)=
    \begin{cases}
        \displaystyle 1-\alpha\,\frac{T_i-T_{\min}}{T_i}, & r(\tau_i) = 1,\\[6pt]
        0, & r(\tau_i) = 0,
    \end{cases}
    \qquad
    T_{\min}=\min_{\tau_j \in \mathcal{T}_{\mathrm{succ}}}|\tau_j|,
    \;\alpha\in(0,1],
    \label{eq:spa}
\end{equation}
where \(T_i = |\tau_i|\) is the length of trajectory \(\tau_i\), and 
\(\mathcal{T}_{\mathrm{succ}}=\{\tau_j \mid r(\tau_j)=1\}\) denotes the set of 
\emph{successful} trajectories for the current problem instance.
Here \(T_{\min}\) is the length of the shortest successful trajectory in \(\mathcal{T}_{\mathrm{succ}}\), 
and \(\alpha\!\in(0,1]\) controls the penalty strength.
In this formulation, shorter sequences are not automatically considered better; 
unsuccessful early terminations still receive a reward of \(0\). 
This adjustment encourages the policy to prefer shorter, 
successful paths without sacrificing the success rate.

\paragraph{Trajectory-level Advantage.}
Given a task, we sample a group of $G$ trajectories $\mathcal{G}=\{\tau_1,\ldots,\tau_G\}$.  
For trajectory $\tau_i$ with $T_i$ valid steps, the SPA reward at each step is given by \eqref{eq:spa}.   
We compute step-wise group-relative advantages by standardizing $R^{\mathrm{SPA}}(s_t,a_t)$:
\[
A_{i,t}\;=\;
\frac{
    R^{\mathrm{SPA}}(s_{i,t},a_{i,t})
    - \mathrm{avg}_{j\in\mathcal{G}} R^{\mathrm{SPA}}(s_{j,t},a_{j,t})
}{
    \mathrm{std}_{j\in\mathcal{G}} R^{\mathrm{SPA}}(s_{j,t},a_{j,t})
},
\qquad t=1,\ldots,T_i.
\]
Finally, the trajectory-level advantage is obtained by averaging over its steps:
\begin{equation}
\label{eq:traj-adv}
A_{\mathrm{traj}}(\tau_i)=\frac{1}{T_i}\sum_{t=1}^{T_i} A_{i,t}.
\end{equation}

\subsubsection{Difficulty–Adaptive Positive Replay (AdaPR)}
Difficulty-Adaptive Positive Replay (AdaPR) strategically retains and reuses challenging, high-value trajectories while blending them with fresh on-policy samples. In sparse-reward mobile environments, successful but difficult rollouts are rare yet highly informative; leveraging them effectively strengthens the learning signal and stabilizes policy updates.
We introduce three key components of AdaPR: buffer construction for high-quality trajectory selection, mixture sampling to balance replay and exploration, and negative rollout pruning to reduce noise.

\vspace{2pt}\paragraph{Buffer construction.}
At iteration $t$, the rollout set is
$\mathcal{T}_t=\{\tau^{(1)},\dots,\tau^{(N)}\}$, collected under
the current policy $\pi_{\theta_t}$.
We compute $A_{\mathrm{traj}}(\tau)$ via \eqref{eq:traj-adv}
and insert the top $\kappa$ trajectories into  
the replay buffer $\mathcal{B}$.

\vspace{2pt}\paragraph{Mixture sampling.}
Each policy update is performed on a mini‑batch of $M$ trajectories obtained
from the mixture distribution
\begin{equation}
    \label{eq:mixture}
    q(\tau)\;=\;\gamma\,p_{\mathcal{B}}(\tau)
              \;+\;(1-\gamma)\,p_{\mathrm{on}}(\tau)
\end{equation}
where $p_{\mathrm{on}}$ is the on‑policy distribution
$\pi_{\theta_t}$ and $p_{\mathcal{B}}$ is the empirical distribution over
$\mathcal{B}$.
To keep the replay contribution under control, at most
$\gamma M$ trajectories with the highest \emph{current}
advantage are drawn from $\mathcal{B}$, preserving on‑policy diversity.

\vspace{2pt}\paragraph{Pruning negative rollouts.}
To further stabilize training, we selectively prune trajectories with the lowest advantages to reduce the probability of introducing noisy samples into the replay buffer. Since high-advantage trajectories, if stored, are expected to be additionally sampled only once on average, we enforce a positive-to-negative trajectory ratio of at most \(1{:}2\) by discarding the lowest-advantage trajectories in descending order of trajectory-level Kullback–Leibler (KL) divergence. The KL divergence is computed between the current policy \(\pi_{\theta_t}\) and the behavior policy that generated the trajectory \(\tau\).


\subsubsection{Failure Curriculum Filtering (FCF)}

To avoid repeatedly sampling tasks that yield zero reward—which wastes computation and hinders the collection of positive advantage data—we propose Failure Curriculum Filtering (FCF). In FCF, any task producing all-zero rewards for two consecutive epochs enters a three-epoch cooldown, during which its sampling probability is reduced according to $w_{\text{task}} = \exp(-f)$, where $f$ is the number of consecutive failure epochs. Tasks with consistently high failure counts are permanently removed from the sampling pool. For stability, failure histories from previous training are retained.

\subsection{Training}
The overall training framework consists of three stages:
\begin{itemize}[leftmargin=*,itemsep=0pt,parsep=0.2em,topsep=0.2em,partopsep=0.0em]
    \item \textbf{Supervised fine-tuning~(SFT):} We curate an instruction-tuning expert dataset covering more than 50 applications and 500k interaction steps. This dataset is used to perform reasoning-free supervised fine-tuning of the base model.
    \item \textbf{\reasoningsft(IRR):} Building upon the supervised stage, we employ the \reasoningsft procedure on the expert dataset to construct a reasoning-oriented fine-tuning corpus. Training on this corpus for an additional 60k steps yields a reasoning warm-up model.
    \item \textbf{\onlinerl:} In the final stage, we perform online sampling from a curated suite of tasks to optimize agent behavior. 
\end{itemize}
For fair comparison with the current state-of-the-art UI-Tars-1.5~\citep{qin2025uitarspioneeringautomatedgui}, we adopt Qwen2.5-VL-7B-Instruct~\citep{Qwen2.5-VL} as our initialization model. In addition, we apply the same training framework to GLM-4.1V-9B-Base~\citep{glmvteam2025glm41vthinkingversatilemultimodalreasoning}, resulting in \textbf{\modelglm}.

}
\section{Experiments}
\begin{table}[t!]
\renewcommand\tabcolsep{15pt}
\centering
\caption{Success rates (\%) of proprietary and open-source models on AndroidWorld and AndroidLab for mobile GUI interaction tasks.}
\vspace{-2mm}
\renewcommand\tabcolsep{6pt}
\renewcommand\arraystretch{1.1}
\label{tab:android_benchmarks}

\begin{tabular}{@{}lccc@{}}
\specialrule{1.0pt}{1.0pt}{1.0pt}
\textbf{Models} & \textbf{\#Params} & {AndroidWorld} & {AndroidLab} \\
\specialrule{1.0pt}{1.0pt}{1.0pt}
\multicolumn{4}{@{}l}{\textit{Proprietary Models}} \\
GPT-4o-2024-11-20~\citep{openai2023gpt4} & - & 34.5 & 31.2 \\
Claude-Sonnet-4-20250514-thinking~\citep{Claude} & - & 41.0 & 40.6 \\
UI-Tars-1.5~\citep{qin2025uitarspioneeringautomatedgui} & - & \underline{64.2} & 38.3 \\
\textsc{AutoGLM}-2024-10~\citep{liu2024autoglmautonomousfoundationagents} & - & -- & 36.2 \\
\midrule[\heavyrulewidth]
\multicolumn{4}{@{}l}{\textit{Open Models}} \\
Qwen2.5-VL-7B-Instruct~\citep{Qwen2.5-VL} & 7B & 27.6 & 10.1 \\
GLM-4.1V-9B-Thinking~\citep{glmvteam2025glm41vthinkingversatilemultimodalreasoning} & 9B & 41.7 & 24.6 \\
UI-Tars-7B~\citep{qin2025uitarspioneeringautomatedgui} & 7B & 33.0 & 32.6 \\
V-Droid~\citep{dai2025advancingmobileguiagents} & 8B & 59.5 & 38.3 \\
UI-Genie-Agent~\citep{xiao2025uigenieselfimprovingapproachiteratively} & 72B & - & \underline{41.2} \\
\midrule[\heavyrulewidth]
\multicolumn{4}{@{}l}{\textit{\method (Ours)}} \\
\quad w/ Qwen2.5-VL-7B & 7B & 72.0 & 42.5 \\
\quad w/ GLM-4.1V-9B-Base  & 9B & \textbf{80.2} & \textbf{53.6} \\
\specialrule{1.0pt}{1.0pt}{1.0pt}
\end{tabular}

\end{table}

\subsection{Experiments Settings}
\paragraph{Datasets and Benchmarks.}
In the two-stage SFT process, we utilize data from human annotations and AndroidControl~\citep{li2024effects}, constructing 97.9k and 23.6k training steps, respectively.  
During the RL stage, we employ interactive training sets from AndroidWorld~\citep{rawles2024androidworlddynamicbenchmarkingenvironment} and AndroidLab~\citep{xu2024androidlabtrainingsystematicbenchmarking}, consisting of 2,000 and 1,103 tasks. These were paired with verifiable rewards and VLM-based reward models (cf. Appendix~\ref{appendix:reward-model}).  
We evaluate on two interactive Mobile benchmarks: AndroidWorld, which includes 116 tasks across 20 apps, and AndroidLab, which comprises 138 tasks across 9 apps. Both benchmarks provide interactive environments.  
We further assess performance on the static dataset AndroidControl~\citep{li2024effects}, which contains 8,444 test samples. Due to space limitations, additional details on the training and test datasets are provided in Appendix~\ref{appendix:data-processing}.

\paragraph{Baselines.}
Our baselines encompass both closed‑ and open‑source agents and models. 
The closed‑source LLMs include GPT‑4o‑2024‑11‑20~\citep{openai2023gpt4} and Claude-Sonnet-4-20250514-thinking~\citep{Claude}, and closed‑source agents UI‑Tars‑1.5~\citep{qin2025uitarspioneeringautomatedgui} and AutoGLM~\citep{liu2024autoglmautonomousfoundationagents}. 
The open‑source VLMs, including Qwen2.5‑VL‑7B-Instruct~\citep{Qwen2.5-VL}, GLM-4.1V-9B-Thinking~\citep{glmvteam2025glm41vthinkingversatilemultimodalreasoning}, UI‑Tars‑7B~\citep{qin2025uitarspioneeringautomatedgui}, V‑Droid~\citep{dai2025advancingmobileguiagents} and UI-Genie-Agent~\citep{xiao2025uigenieselfimprovingapproachiteratively}, 
are used. 



\subsection{Main Results}

We evaluate \method by using Qwen2.5-VL-7B and GLM-4.1V-9B-Base as backbones on online interactive benchmarks, AndroidWorld and AndroidLab. 
As shown in Table~\ref{tab:android_benchmarks}, \method significantly outperforms previous results. 
With Qwen2.5-VL-7B as the backbone, \method achieves \qwaw~ on AndroidWorld and \qwal~ on AndroidLab, outperforming previous state-of-the-art methods. 
By using GLM-4.1V-9B as the backbone, the performance of \method further improves to \glmaw~ on AndroidWorld and \glmal~ on AndroidLab, achieving the best results across all models.




\subsection{Ablation Study}

\newcommand{\secondrowratio}{0.90}  
\newcommand{\bleftratio}{0.55}      
\newcommand{\brightratio}{0.4}     
\newcommand{\bcgapratio}{0.01}      

\begin{figure*}[t]
\centering
\begin{subfigure}[c]{0.85\textwidth}
  \centering
    \begin{tabular}{@{}lll@{}}
      \toprule
      \textbf{Models} & {AndroidWorld} & {AndroidLab} \\
      \midrule
      Qwen2.5-VL-7B-Instruct~\citep{Qwen2.5-VL} & 27.6 & 10.1 \\
      \hdashline
      \quad + Reasoning-Free SFT & 50.2\textsubscript{\scriptsize\textcolor{darkred}{+22.6}} & 36.9\textsubscript{\scriptsize\textcolor{darkred}{+26.8}} \\
      \quad \quad + Reasoning SFT & 56.8\textsubscript{\scriptsize\textcolor{darkred}{+6.6}} & 38.7\textsubscript{\scriptsize\textcolor{darkred}{+1.8}} \\
      \quad \quad \quad + \onlinerl~(\method-7B) & 72.0\textsubscript{\scriptsize\textcolor{darkred}{+15.2}} & 42.5\textsubscript{\scriptsize\textcolor{darkred}{+3.8}} \\
      \midrule
    GLM-4.1V-9B-Base~\citep{glmvteam2025glm41vthinkingversatilemultimodalreasoning} & 7.7  & 10.1 \\
    \hdashline
    \quad + Reasoning-Free SFT & 48.1\textsubscript{\scriptsize\textcolor{darkred}{+40.4}} & 42.7\textsubscript{\scriptsize\textcolor{darkred}{+32.6}} \\
    \quad \quad + Reasoning SFT & 66.2\textsubscript{\scriptsize\textcolor{darkred}{+18.1}} & 45.0\textsubscript{\scriptsize\textcolor{darkred}{+2.3}} \\
    \quad \quad \quad + \onlinerl~(\method-9B) & 80.2\textsubscript{\scriptsize\textcolor{darkred}{+14.0}} & 53.6\textsubscript{\scriptsize\textcolor{darkred}{+8.6}} \\

      \bottomrule
    \end{tabular}
  \caption{Improvements of task success rates by incrementally applying Reasoning-Free SFT, Reasoning SFT, and \onlinerl to the base models, respectively. }
  \label{fig:framework_ablation_table}

\end{subfigure}

\begin{center}
\begin{minipage}{\secondrowratio\textwidth}
  \centering
  \begin{subfigure}[c]{\bleftratio\linewidth}   
    \begin{minipage}[c][\ablationboxht][c]{\linewidth}
      \centering
      \includegraphics[width=0.9\linewidth]{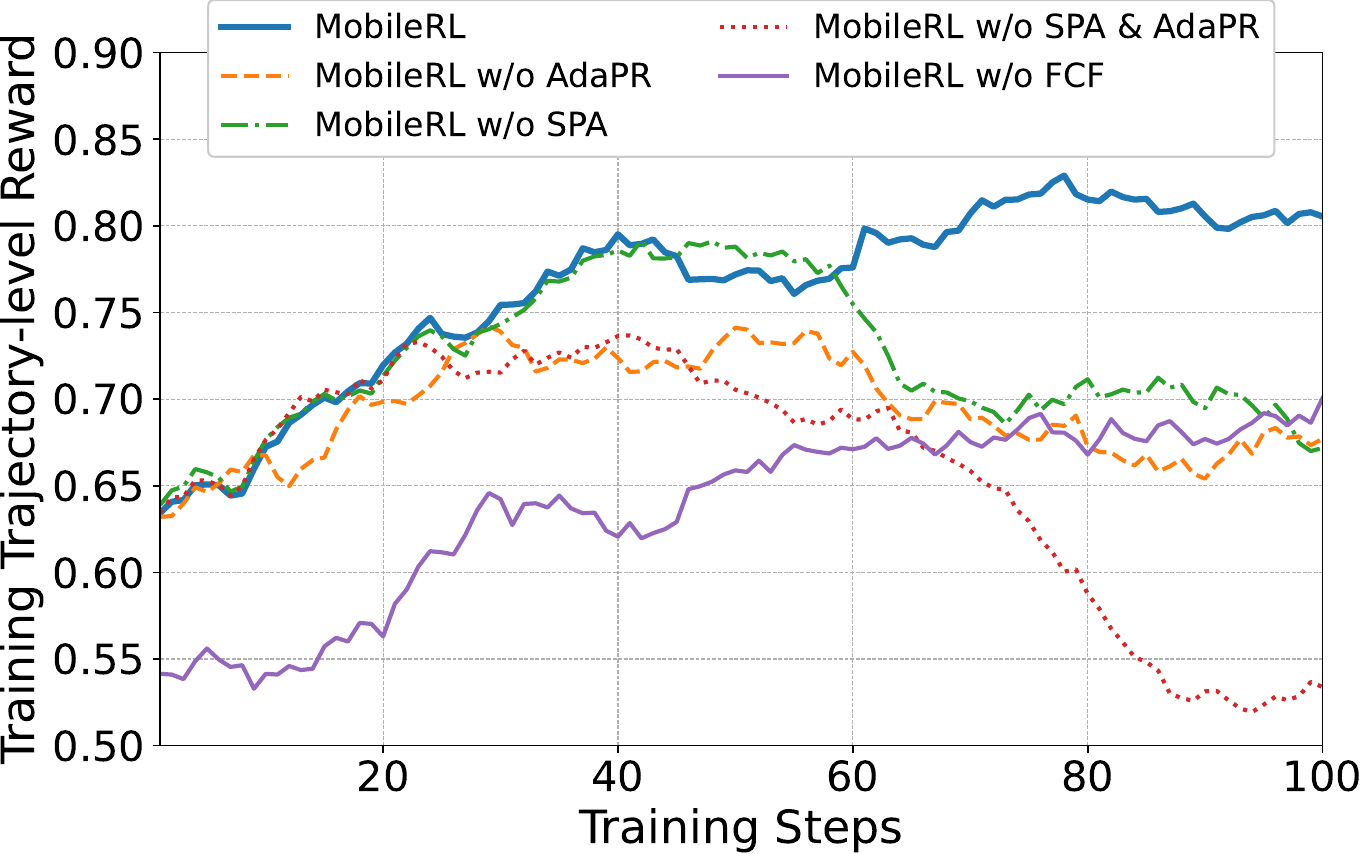}
    \end{minipage}
    \vspace{0.8em}
    \caption{Training trajectory-level rewards from the AndroidWorld environment with respect to training steps.}
    \label{fig:ablation_curve}
  \end{subfigure}
  \hspace{\bcgapratio\linewidth}                
  \begin{subfigure}[c]{\brightratio\linewidth}  
    \begin{minipage}[c][\ablationboxht][c]{\linewidth}
      \centering
        \begin{tabular}{@{}lc@{}}
          \toprule
          \textbf{Models} & {AndroidWorld} \\
          \midrule
          \method & \textbf{71.1} \\
          \quad w/o AdaPR & 63.6 \\
          \quad w/o SPA & 69.1 \\
          \quad w/o AdaPR\&SPA & 58.5 \\
          \quad w/o FCF & 64.8 \\
          \midrule
          \quad w/o \onlinerl & 56.8 \\
          \bottomrule
        \end{tabular}
    \end{minipage}
    \vspace{0.8em}
    \caption{Test performance on the AndroidWorld test set under different variants.}
    \label{fig:ablation_table}
  \end{subfigure}
\end{minipage}
\end{center}

\caption{
  Ablation studies of the \method framework and its \onlinerl algorithm.  
  We use the Reasoning SFT model with Qwen2.5-VL-7B-Instruct backbone for the ablation of the \onlinerl algorithm.
  All test set results are averaged over three runs to mitigate the impact of randomness.
}
\label{fig:combined_framework_ablation}
\vspace{-5mm}
\end{figure*}

To evaluate the contributions of the \method framework and the components of the \onlinerl algorithm, we start with two base models and progressively apply Reasoning-Free SFT, Reasoning SFT, and \onlinerl. 
Then, using the Qwen2.5-VL-7B-Instruct model trained with Reasoning SFT as the initialization point and on the AndroidWorld training set, we conduct an analysis of the impact of each component of \onlinerl---AdaPR, SPA, and FCF.

\paragraph{\method Ablation.}
We summarize stage-wise gains in success rate (SR) in Table~\ref{fig:framework_ablation_table}. 
For Qwen2.5-VL-7B, the combined improvements of \method are +44.4\% on AndroidWorld and +32.4\% on AndroidLab; 
For GLM-4.1V-9B, the overall gains are +72.5\% and +43.5\%. 
Overall, Reasoning-Free SFT delivers a strong initial lift, and Reasoning SFT offers additional improvements. 
Building upon the strong foundation established by the preceding stages, the \onlinerl stage further augments the final performance, achieving an improvement exceeding 10\% on the AndroidWorld dataset and over 5\% on the AndroidLab test set.

\paragraph{\onlinerl Ablation.}
The design of \onlinerl covers SPA, AdaPR, and FCF. 
The ablations are performed on four settings: 
(i) {\method w/o AdaPR} (no replay), 
(ii) {\method w/o SPA} (no reward shaping), 
(iii) {\method w/o AdaPR \& SPA} (neither), 
and (iv) {\method w/o FCF} (uniform sampling). 

We report on-policy trajectory reward curves during training (excluding replayed trajectories) in Figure~\ref{fig:ablation_curve} and the success rates on the AndroidWorld in Table~\ref{fig:ablation_table}.
To avoid bias from the AndroidLab reward model, we use only AndroidWorld in this study. 
Each component of \onlinerl contributes to improving the performance of \method. 
Specifically, we have the following observations: 
\begin{itemize}[leftmargin=*,itemsep=0pt,parsep=0.2em,topsep=0.2em,partopsep=0.0em]

\item \textbf{FCF under constraints.} 
With a 100-step budget ($>40$ hours), FCF plays a key role in filtering.  Removing it biases early sampling toward overly hard tasks, yielding many negatives and a lower reward ceiling. 
The w/o FCF curve keeps rising, suggesting stronger results with more efficient pipelines or simulators.

\item \textbf{FCF only (w/o AdaPR \& SPA).} 
Training is initially stable but collapses after about 30 steps, indicating that AdaPR and SPA are necessary for stabilizing the training.

\item \textbf{Effect of AdaPR.} 
After about seven steps (once the replay buffer is populated), the gap between w/o AdaPR and the full \method method grows, showing the benefit of replay. 

\item \textbf{Effect of SPA.} 
Noticeable gains of SPA appear after roughly 60 steps, likely because the lack of step-wise control leads to overly long trajectories.
\end{itemize}



\paragraph{Is Reasoning-Free SFT Still Necessary?}

\label{exp:reasoning-free}

We apply supervised fine-tuning on the expert dataset without reasoning traces, which we term \textit{Reasoning-Free SFT}. 
This raises a key question: \textit{Is fine-tuning with expert data that lacks reasoning still beneficial?}

We compare \method (Reasoning-Free SFT + Reasoning SFT + \onlinerl) with \method without Reasoning-Free SFT  in Figure~\ref{fig:ablation-sft}.
Interestingly, our experiments show that incorporating Reasoning-Free SFT consistently improves performance on AndroidWorld. 
It suggests that without explicit reasoning traces, (Reasoning-Free) SFT contributes to enhancing final results for \method.




\begin{table}[t]
\centering
\begin{minipage}{0.52\textwidth}
    \centering
    \caption{Results on the AndroidControl~\citep{li2024effects}. Note that data from AndroidControl was not included during the RL stage.}
    \label{tab:android_control_sr}
    \resizebox{\linewidth}{!}{%
        \begin{tabular}{lcccccc}
\toprule
\multirow{2}{*}{Model} 
  & \multicolumn{3}{c}{AndroidControl-Low} 
  & \multicolumn{3}{c}{AndroidControl-High} \\
  \cmidrule(lr){2-4} \cmidrule(lr){5-7}
  & Type & Grounding & SR & Type & Grounding & SR \\
\midrule
UI-Tars-7B                 & 98.0 & 89.3 & 90.8 & 83.7 & 80.5 & 72.5 \\
UI-Tars-72B                & 98.1 & 90.8 & 91.3 & 85.2 & 81.5 & 74.7 \\
UI-Genie-7B                & 98.1 & 89.3 & 94.3 & 83.5 & 82.9 & 74.5 \\
UI-Genie-72B               & 98.3 & 95.4 & 94.8 & 84.9 & \textbf{86.3} & 77.0 \\
\midrule
\multicolumn{7}{@{}l}{\textit{\method-7B}} \\ 
\quad w/o RL               & 96.7 & 93.7 & 91.9 & 87.0 & 73.6 & 72.8 \\ 
\quad w/  RL               & 96.9 & 92.4 & 91.4 & 86.3 & 71.5 & 71.3 \\ 
\multicolumn{7}{@{}l}{\textit{\method-9B}} \\ 
\quad w/o RL               & \textbf{98.6} & \textbf{95.8} & \textbf{95.9} & \textbf{89.8} & 78.1 & \textbf{77.5} \\ 
\quad w/  RL               & 97.9 & 94.6 & 94.3 & 87.0 & 74.5 & 73.9 \\ 
\bottomrule

\end{tabular}
    }
\end{minipage}%
\hfill
\begin{minipage}{0.45\textwidth}
    \centering
    \includegraphics[width=\linewidth]{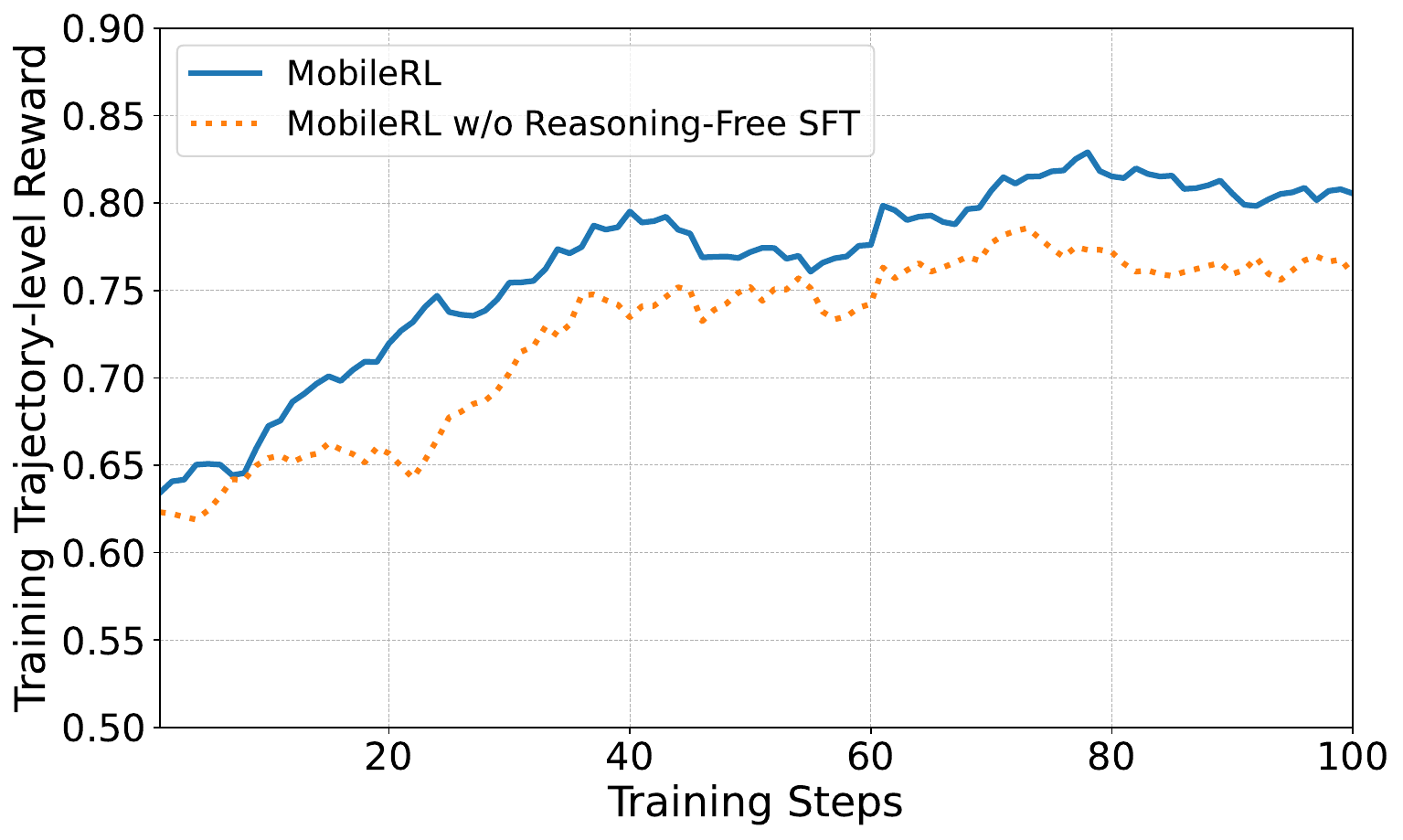}
    \caption{Effect of reasoning-free SFT evaluated on AndroidWorld.}
    \label{fig:ablation-sft}
\end{minipage}
\end{table}

\paragraph{Performance on Offline Dataset.}
As shown in Appendix~\ref{appendix:data-processing}, Reasoning-Free and Reasoning SFT stages include training data from AndroidControl, which has been converted into MobileRL format. In Table~\ref{tab:android_control_sr}, we present the results of MobileRL on the AndroidControl test set. The SR score of MobileRL-9B w/o RL version surpasses all previous models, achieving the state of the art, while the RL version largely maintains the score.

\begin{figure}[t]
    \begin{minipage}{0.43\textwidth}
        \centering
        \includegraphics[width=\textwidth]{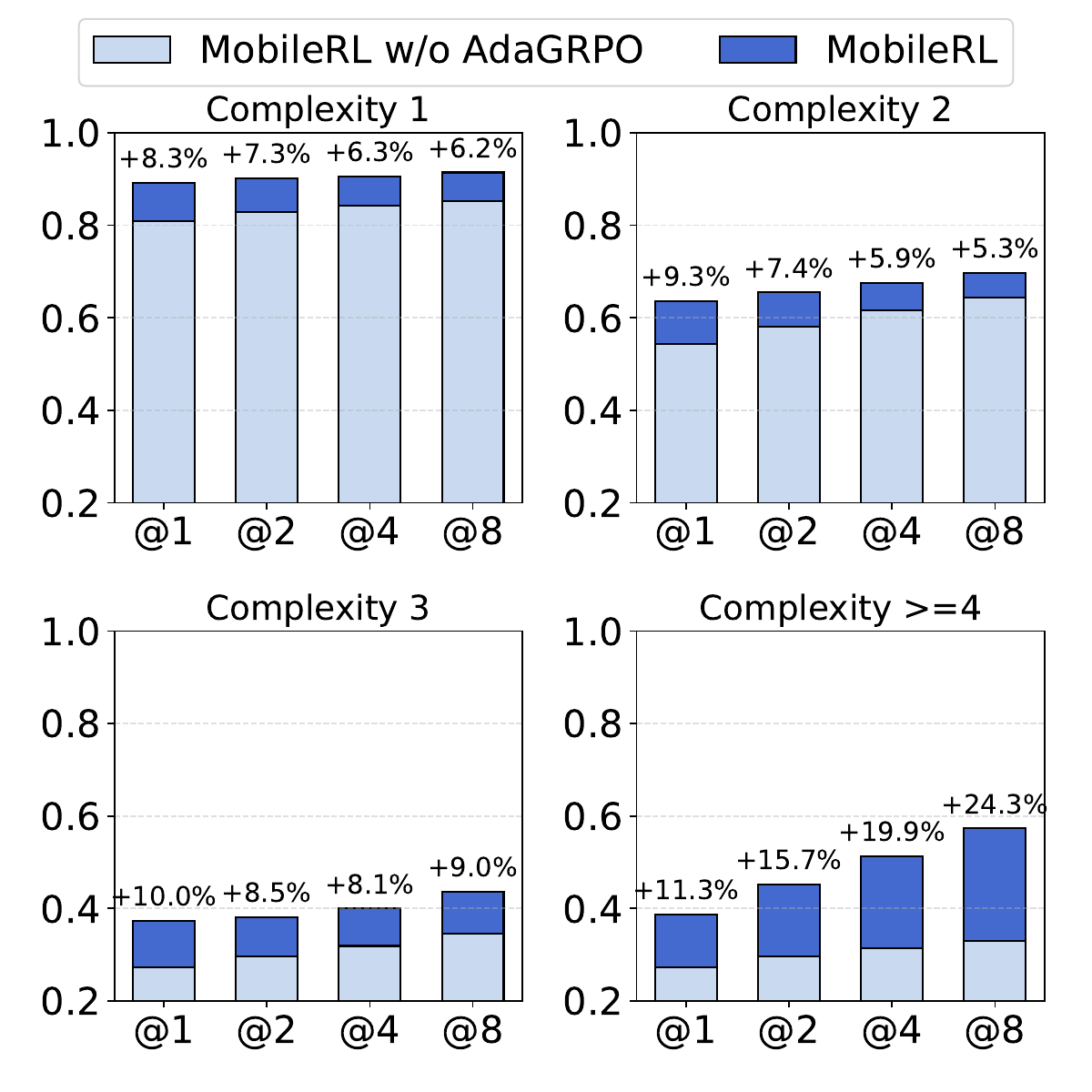}
        \caption{Pass@$k$ on AndroidWorld by task complexity (\cite{rawles2024androidworlddynamicbenchmarkingenvironment}) levels. Pass@$k$ is the fraction of tasks solved within the top-$k$ attempts.}
        \label{fig:passk_complexity}
    \end{minipage}
    \hfill
    \begin{minipage}{0.55\textwidth}
        \centering
        \includegraphics[width=\textwidth]{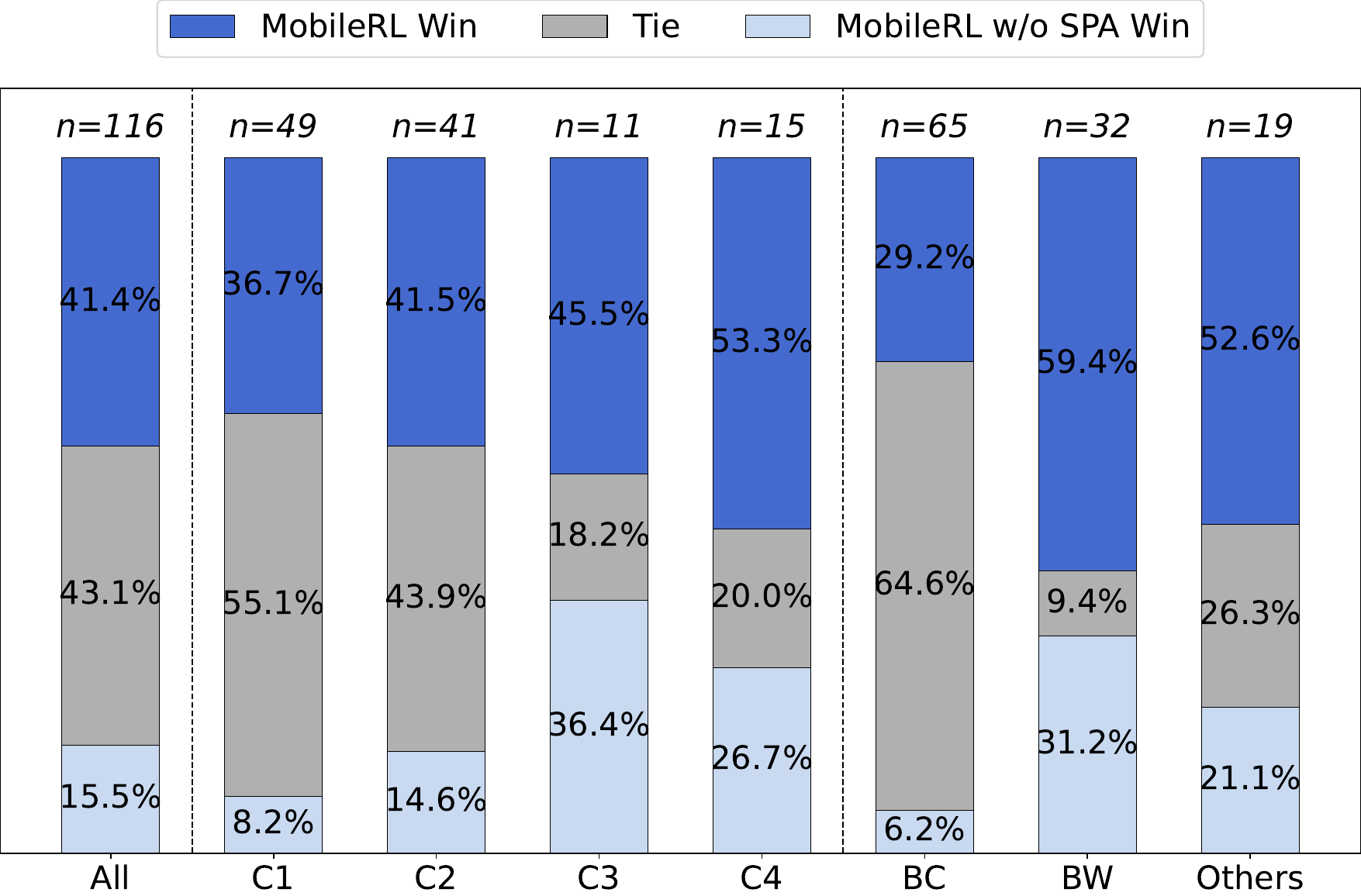}
        \caption{Win rate of \method vs.\ \method\ w/o SPA, where a \emph{win} means completing a task with fewer steps. 
        $n$ denotes the number of task templates per category. 
        Categories: \emph{All} (all templates); \emph{C1--C4} (complexity levels 1--4); \emph{BC}/\emph{BW} (both methods correct/wrong); \emph{Others} (exactly one method correct).}
        \label{fig:win}
    \end{minipage}
\end{figure}

\paragraph{Success Rates by Task Complexity.}
We divide the AndroidWorld test set by rounded-up \emph{Complexity}~\citep{rawles2024androidworlddynamicbenchmarkingenvironment}: complexity level 1 ($\leq$10 steps), level 2 (11--20), level 3 (21--30), and level 4+ ($>$30). 
We run eight test trials at temperature $1.0$ and report pass@1/2/4/8 in Figure~\ref{fig:passk_complexity}.
\method yields consistent gains at all complexity levels. 
Notably, post-RL pass@1 exceeds pre-RL pass@8, indicating substantial effectiveness. 
Consistent with AdaPR’s design for heavy-tailed difficulty, larger gains occur on high-complexity tasks.



\paragraph{Impact of SPA on Step Efficiency.}
\label{exp:spa}
Although SPA has the smallest impact on overall accuracy in the ablation, its effect on step efficiency is clear. As shown in Figure~\ref{fig:win}, partitioning tasks by complexity reveals that \method with SPA consistently completes tasks in fewer steps across all difficulty levels. Moreover, when we compare cases where both models are correct (BC), both are wrong (BW), and those where only one is correct, \method with SPA more frequently yields shorter trajectories in every group.

\section{Related Work}

\paragraph{Mobile GUI Agents.}
Recent work leverages powerful language models to build agents that operate real PCs and phones~\citep{agashe2025agents2compositionalgeneralistspecialist, qin2025uitarspioneeringautomatedgui, lai2025computerrlscalingendtoendonline}, including Android agents that perceive GUIs and act via taps, swipes, and text~\citep{toyama2021androidenv,xu2024androidlabtrainingsystematicbenchmarking}. To improve action prediction and learning, frameworks explore multimodal exploration~\citep{yang2023appagent}, modular reasoning~\citep{lai2025androidgenbuildingandroidlanguage}, verifier-driven control~\citep{dai2025advancingmobileguiagents}, and small-LM code-based execution~\citep{wen2025autodroidv2boostingslmbasedgui}. Yet many systems still rely on offline RL or single-turn data: DigiRL uses offline demonstrations~\citep{bai2024digirltraininginthewilddevicecontrol}; U1-R1 trains on single-step episodes~\citep{lu2025uir1enhancingefficientaction}; and UI-Tars applies DPO in an offline regime~\citep{qin2025uitarspioneeringautomatedgui}, leaving online, multi-turn RL for adaptive mobile agents unexplored. 
Studies emphasize realistic applications: AppAgent~\citep{yang2023appagent}  evaluates closed-source models on real-world apps, while A3~\citep{chai2025a3androidagentarena} offers a realistic app suite with an autonomous evaluation protocol that reduces human effort.

\paragraph{Benchmarks for Mobile Agents.}
Benchmarking generally follows two tracks. Static or replay-based settings—AndroidControl~\citep{li2024effects}, Android in the Wild~\citep{rawles2023aitw}, MobileAgentBench~\citep{wang2024mobileagentbench}, and Mobile-Bench~\citep{deng2024mobile} offer plenty of tasks and trajectories. These benchmarks fall short for real-world evaluation, as fixed trajectories and screens limit agents’ ability to handle uncertainty and exploration. Interactive emulator environments—AndroidWorld~\citep{rawles2024androidworlddynamicbenchmarkingenvironment}, AndroidLab~\citep{xu2024androidlabtrainingsystematicbenchmarking}, and B-MOCA~\citep{lee2024benchmarkingmobiledevicecontrol}—span diverse apps and realistic tasks, yet remain challenging for current agents. Current mobile GUI benchmarks are limited: they lack asynchronous, parallel VM interaction for scalable training.
To the best of our knowledge, all public mobile GUI benchmarks to date target the Android operating system.
\section{Conclusion}

In this work, we present \method, an agentic RL framework that advances mobile GUI agents. 
It achieves this by combining staged initialization with an adaptive reinforcement learning algorithm (\onlinerl). 
Training begins with reasoning-free SFT on large-scale expert demonstrations, followed by a reasoning SFT stage that adds intermediate rationales and reduces cold-start exploration costs.
Building on this, we introduce Difficulty–\textbf{ADA}ptive GRPO (\onlinerl), which enhances GRPO with shortest-path reward adjustment, adaptive positive replay, and failure curriculum filtering. 
These three strategies improve sample efficiency and guide policies toward more accurate and efficient task completion. 
Experiments on AndroidWorld and AndroidLab demonstrate that \method with open models significantly outperforms both open-source and proprietary baselines. 


\bibliography{iclr2026_conference}
\bibliographystyle{iclr2026_conference}

\appendix
\section{Case Study}



\begin{figure*}[t]
    \centering
    \begin{subfigure}[t]{0.30\textwidth}
        \centering
        \includegraphics[width=\linewidth]{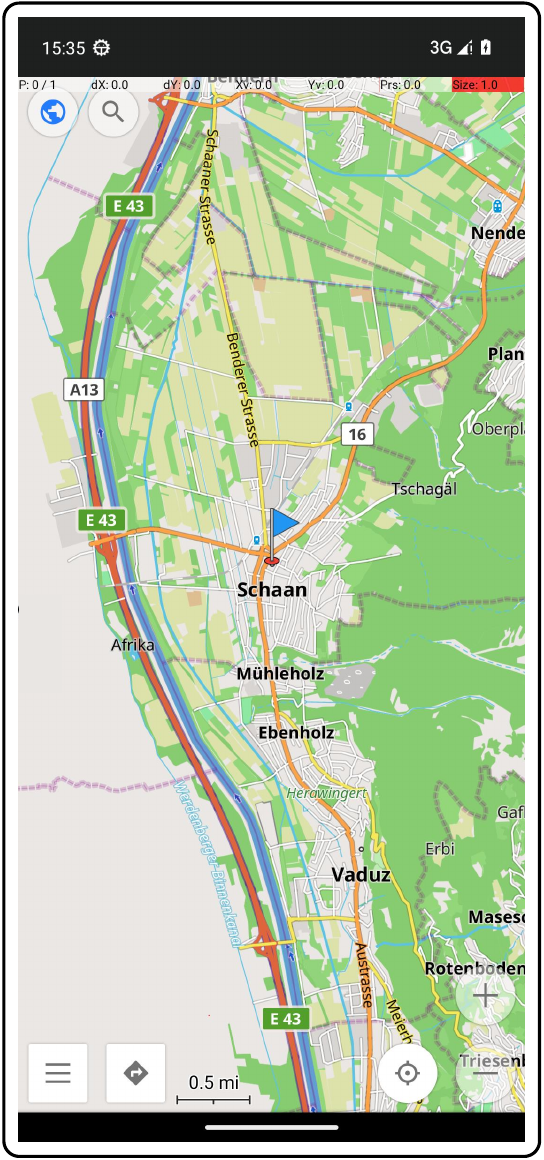}
        \caption{Add a location marker for 47.16, 9.51 in the OsmAnd maps app.}
        \label{fig:sub1}
    \end{subfigure}\hfill
    \begin{subfigure}[t]{0.30\textwidth}
        \centering
        \includegraphics[width=\linewidth]{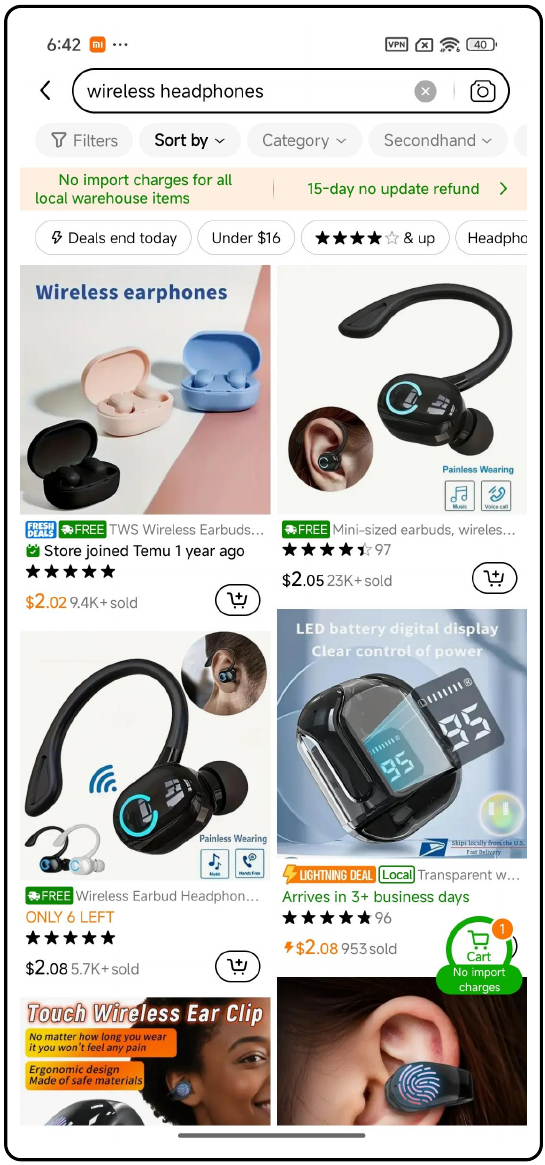}
        \caption{Search for wireless headphones in temu, and sort by price low to high.}
        \label{fig:sub3}
    \end{subfigure}\hfill
    \begin{subfigure}[t]{0.30\textwidth}
        \centering
        \includegraphics[width=\linewidth]{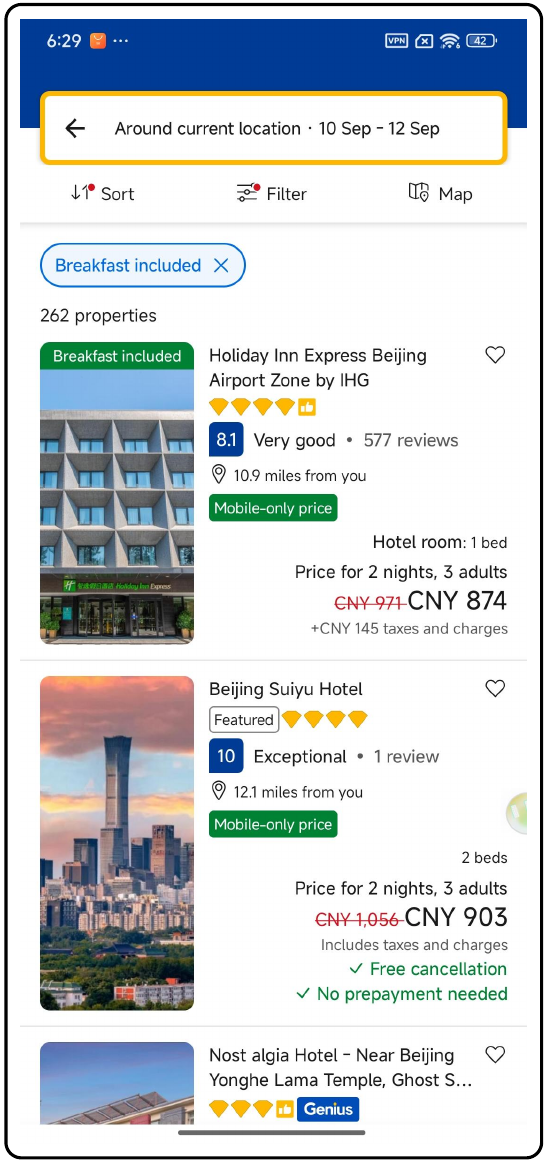}
        \caption{Search for hotels on Booking, check-in date is 09-10, check-out date is 09-12, sort by prices.}
        \label{fig:sub4}
    \end{subfigure}
    \caption{Example mobile tasks finished by our agent. Our agent can automatically perform tasks according to human instructions in academic benchmarks and real-world applications.}
    \label{fig:all}
\end{figure*}

\begin{figure}[t]
    \centering
    \includegraphics[width=\textwidth]{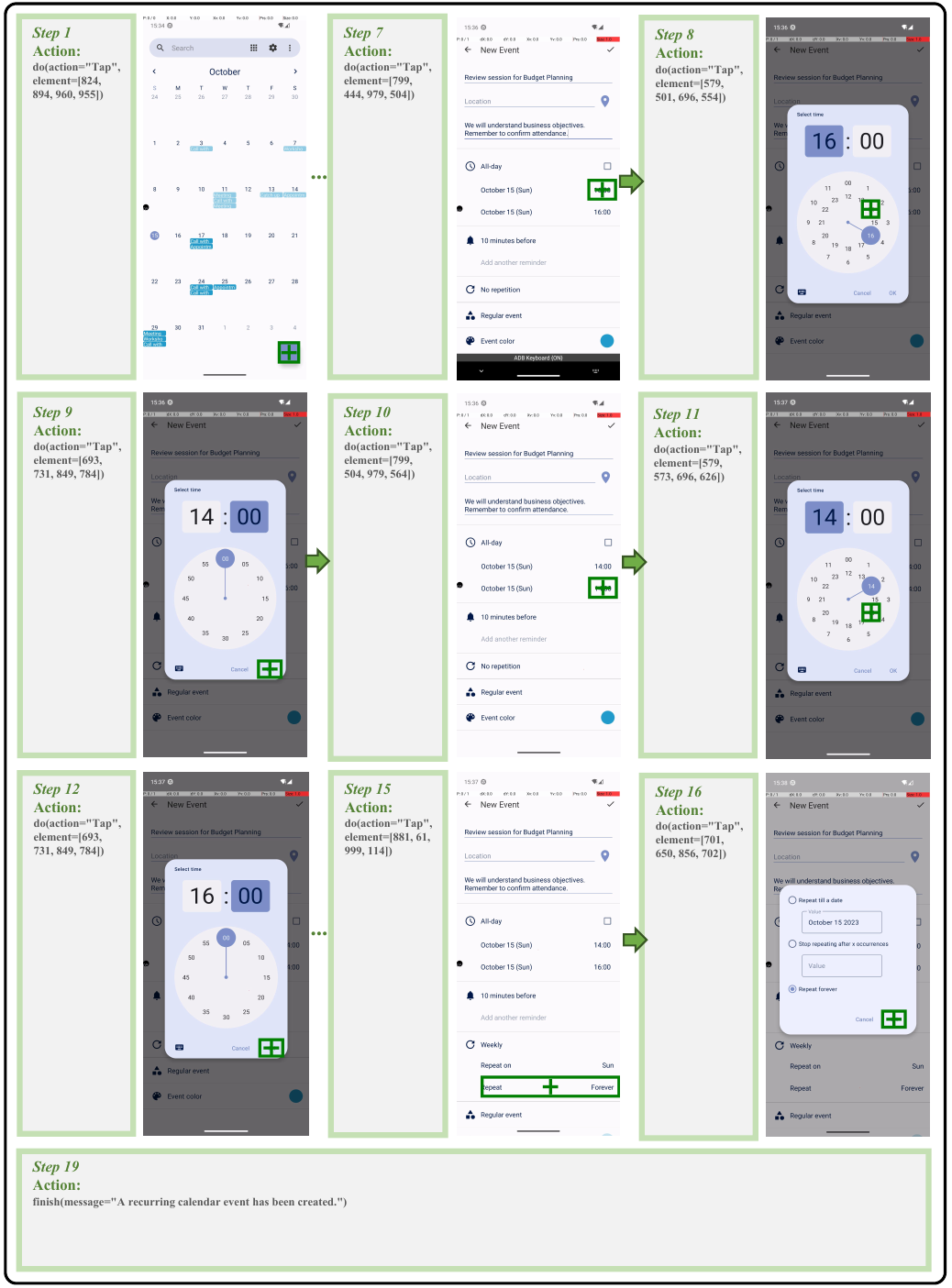}
    \caption{The Reasoning-Free SFT agent, where the expert data does not include the reasoning part, in the \texttt{SimpleCalendarAddRepeatingEvent} task in AndroidWorld. Because of space constraints, we only present the most critical and notable steps for comparison.}
    \label{fig:base-case}
\end{figure}

\begin{figure}[t]
    \centering
    \includegraphics[width=\textwidth]{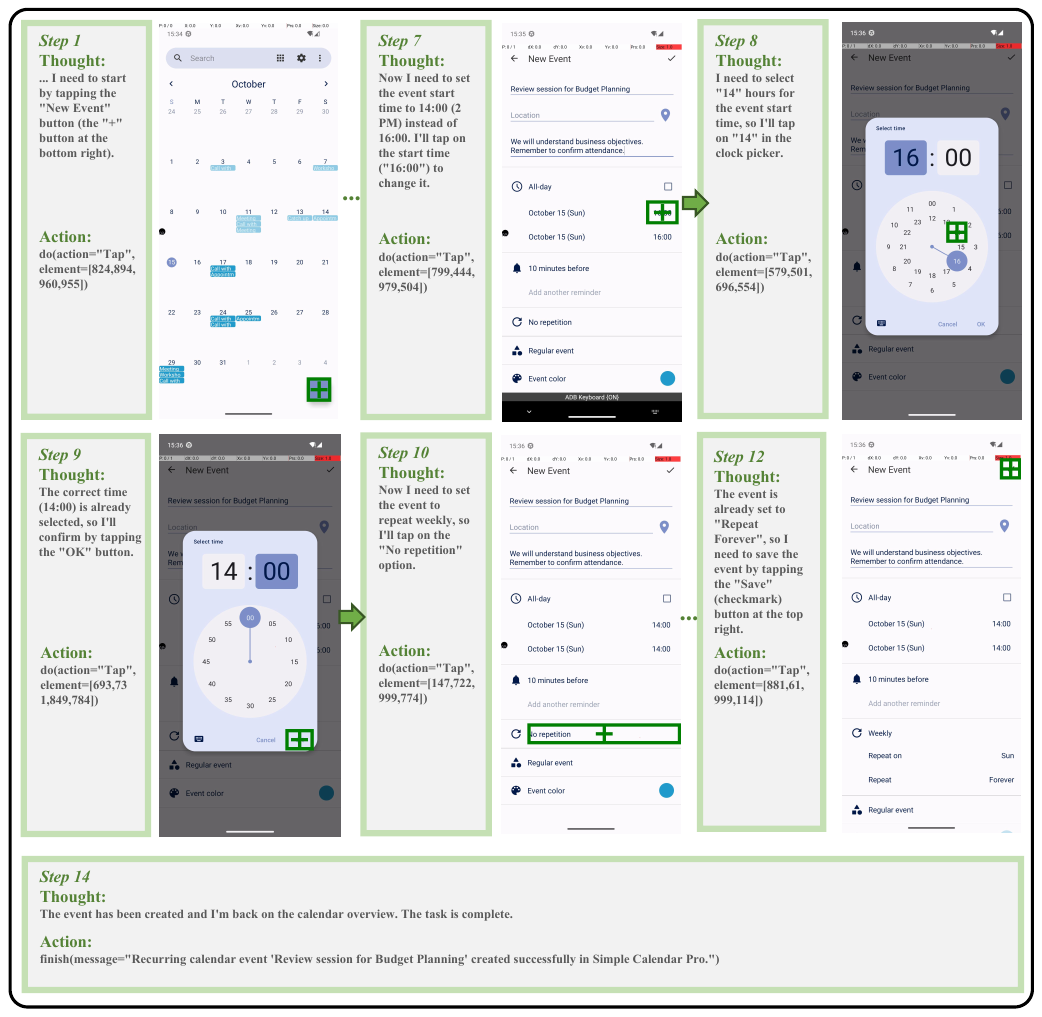}
    \caption{Reasoning SFT agent in the \texttt{SimpleCalendarAddRepeatingEvent} task in AndroidWorld. Because of space constraints, we only present the most critical and notable steps for comparison.}
    \label{fig:sft-case}
\end{figure}

\begin{figure}[t]
    \centering
    \includegraphics[width=\textwidth]{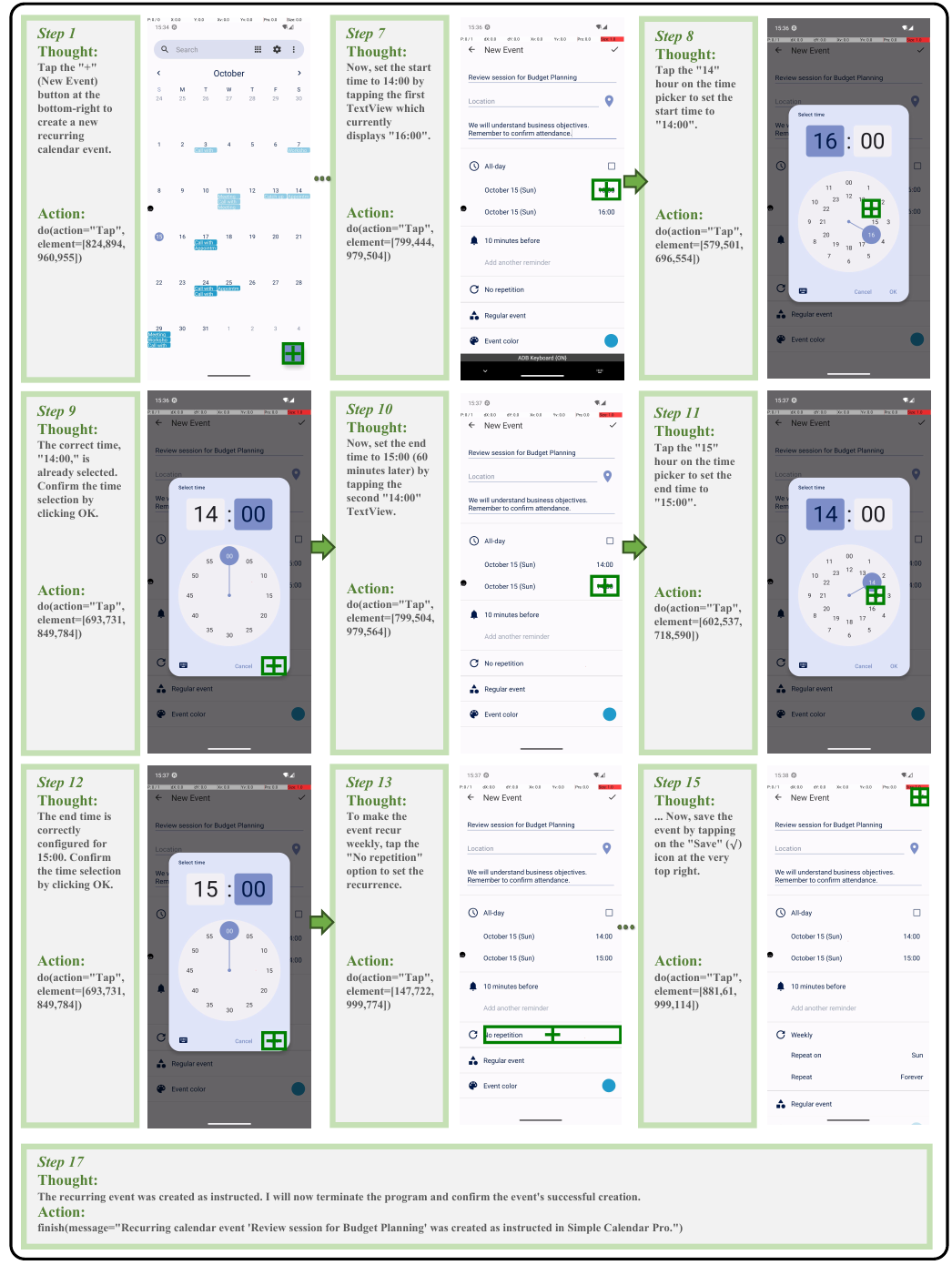}
    \caption{\method in the \texttt{SimpleCalendarAddRepeatingEvent} task in AndroidWorld. Because of space constraints, we only present the most critical and notable steps for comparison.}
    \label{fig:rl-case}
\end{figure}




We present a case study from AndroidWorld evaluating three agents: the Reasoning-Free SFT agent, Reasoning SFT agent, and \method agent. We choose \texttt{SimpleCalendarAddRepeatingEvent}, which requires creating a recurring event titled ``Review session for Budget Planning,'' starting on 2023-10-15 at 14:00, lasting 60 minutes, repeating weekly without end, and including the description: ``We will understand business objectives. Remember to confirm attendance.''

All agents successfully configured the title and basic settings in the initial steps. Their performance diverged in subsequent steps. The Reasoning-Free SFT agent made timing errors (setting 16:00 instead of 15:00) and executed redundant checks, revealing weak task understanding (Figure~\ref{fig:base-case}). The Reasoning SFT agent skipped an adjustment step, yielding an incorrect event duration (Figure~\ref{fig:sft-case}). By contrast, the full \method agent completed the task accurately and efficiently, satisfying all requirements without redundant operations (Figure~\ref{fig:rl-case}).

\section{Training Details}
\label{app:training details}
\subsection{Hyperparameters}
\label{sec:hyperparameters}

In both the Reasoning-Free SFT and Reasoning SFT stages, we conducted fine-tuning for two epochs, employing a cosine-decayed learning rate scheduled from $1 \times 10^{-5}$ to $1 \times 10^{-6}$. The training was implemented using \texttt{Llama-Factory}, with packing mode enabled to accelerate training efficiency. All images were used at their full resolution without compression.

In the RL stage, we extend the GRPO framework from \texttt{Verl} with our customized functionalities. The main hyperparameters are summarized in Table~\ref{tab:hyperparams}.

\begin{table}[h!]
\centering
\caption{Summary of Main Hyperparameters}
\label{tab:hyperparams}
\begin{tabular}{lll}
\toprule
Component & Hyperparameter & Value \\
\midrule
Data & Max Prompt Length & 16384 \\
Data & Max Response Length & 4096 \\
Data & Train Batch Size & 256 \\
Data & Validation Batch Size & 256 \\
\midrule
Actor / Policy & Strategy (Parallelism) &  FSDP \\
Actor / Policy & PPO Micro Batch Size/GPU & 4 \\
Actor / Policy & Learning Rate (LR) & 1e-6 \\
Actor / Policy & Gradient Clipping & 1.0 \\
Actor / Policy & Clip Ratio & 0.2 \\
Actor / Policy & PPO Epochs & 1 \\
\midrule
Rollout \& Sampling & Sampling Temperature & 1.0 \\
Rollout \& Sampling & Max New Tokens & 4096 \\
Rollout \& Sampling & Number of Samples (n) & 16 \\
Rollout \& Sampling & Max Turns & 50 \\
Rollout \& Sampling & Max Pixels & 5000000 \\
Rollout \& Sampling & Min Pixels & 65536 \\
\midrule
Algorithm & KL Loss Coefficient/ $\beta$ & 0.001 \\
Algorithm & SPA/ $\alpha$ & 1.0 \\
Algorithm & \replaybuffer/ Replay Buffer Size & 256 \\
Algorithm & \replaybuffer/ $\gamma$ & 1.0 \\
Algorithm & \replaybuffer/ $\kappa$ & 0.25 \\
\bottomrule
\end{tabular}
\end{table}

\begin{table*}[t]
  \centering
  \caption{Action Space for Mobile GUI Interaction. We utilize a merging action space from AndroidLab~\citep{xu2024androidlabtrainingsystematicbenchmarking} and AndroidControl~\cite{li2024effects}, which represents screen positions with bounding boxes aligned to XML data. 
  }
  \label{tab:actions}
  \begin{tabularx}{\textwidth}{@{}l l X@{}}
    \toprule
    \textbf{Action}   & \textbf{Parameters}                                                                          & \textbf{Description}                                         \\
    \midrule
    Tap               & \texttt{element=[x1,y1,x2,y2]}                                                               & Tap at the rectangle defined by top‐left (x1,y1) and bottom‐right (x2,y2).  \\
    Type              & \texttt{text=\{string\}}                                                                     & Enter the given string into the focused input field.        \\
    Swipe             & \begin{tabular}[t]{@{}l@{}} \texttt{direction=\{up/down/left/right\}}\\ \texttt{dist=\{short/medium/long\}}\\ \texttt{element=[x1,y1,x2,y2] (optional)} \end{tabular}
                      & Swipe in the given direction over the specified distance.  
                      Optionally constrain to the rectangle element.            \\
    Long Press        & \texttt{element=[x1,y1,x2,y2]}                                                               & Press and hold on the given rectangle area.                 \\
    Launch            & \texttt{app=\{AppName\}}                                                                     & Launch the named application.                               \\
    Back              & none                                                                                         & Press the system Back button.                               \\
    Home              & none                                                                                         & Press the system Home button.                               \\    
    Wait              & none                                                                                         & Wait for three seconds.                               \\
    Finish            & \texttt{message=\{string\} (optional)}                                                       & End the session with an optional message.              \\
    \bottomrule
  \end{tabularx}
\end{table*}

\subsection{Action Space.}
We design a series of actions based on AndroidLab~\citep{xu2024androidlabtrainingsystematicbenchmarking}, which supports a concise set of actions for GUI interaction, as described in Table~\ref{tab:actions}.

\subsection{Implement Details}

\paragraph{Pruning Negative Trajectories}

To stabilize training, we prune trajectories with the lowest advantages, reducing noisy samples in the replay buffer. High-advantage trajectories, if stored, are expected to be sampled only once on average. Thus, we maintain a maximum positive-to-negative trajectory ratio of 1:2 by randomly discarding the lowest-advantage trajectories.

\begin{figure}[t]
    \centering
    \includegraphics[width=0.5\textwidth]{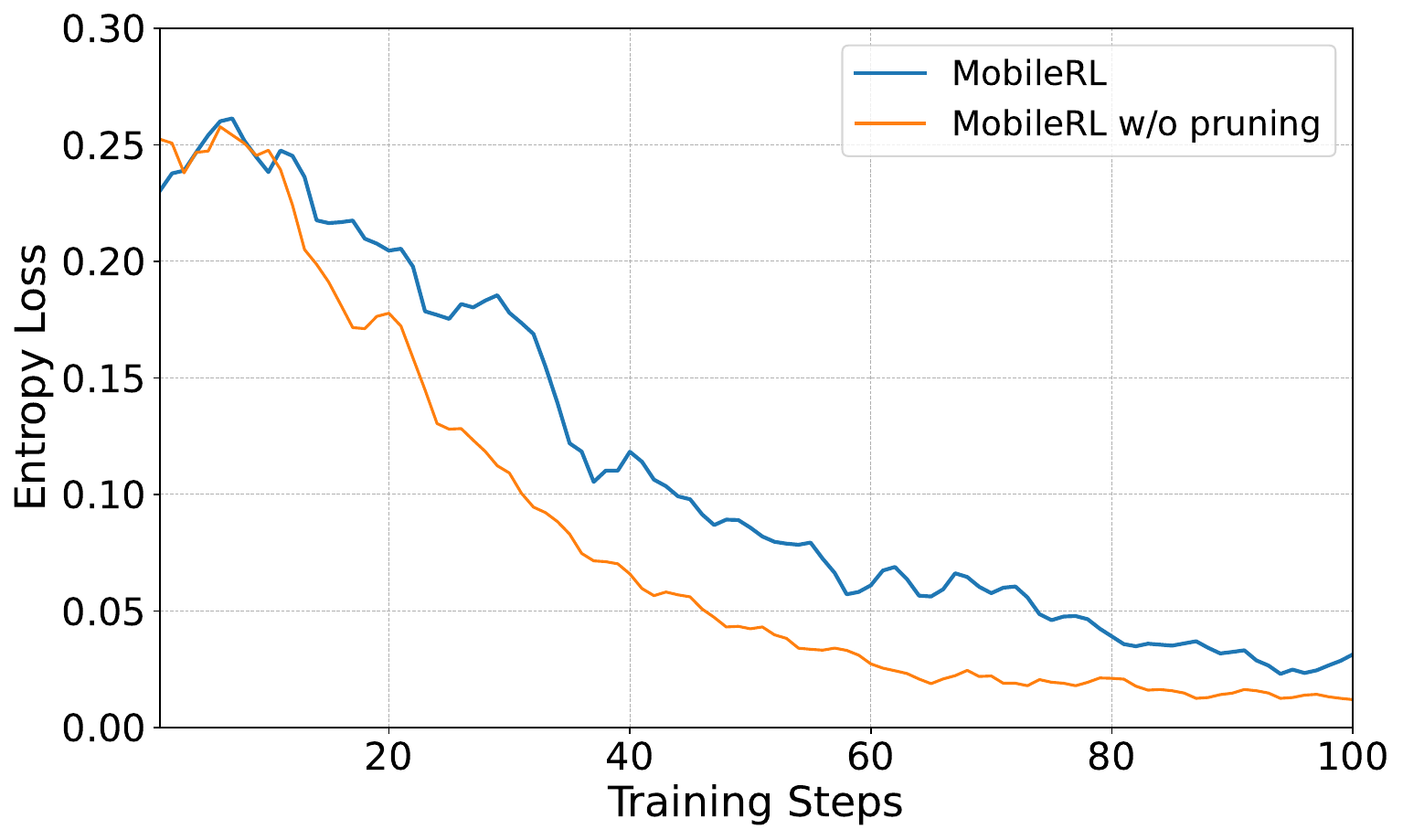}
    \caption{Effect of the pruning-negative strategy evaluated on AndroidWorld.}
    \label{fig:mute_vs_entropy_loss}
\end{figure}

We further study a pruning strategy that discards overly frequent erroneous trajectories from the reinforcement learning buffer before each update. 
As depicted in Figure~\ref{fig:mute_vs_entropy_loss}, this filtering keeps the policy entropy consistently higher during training. 
By pruning trajectories whose advantages remain persistently negative, the agent avoids being driven by detrimental gradients; probability mass is instead spread over a broader action space, fostering exploration and delaying premature convergence. This pruning strategy ultimately leads to more robust policy learning.

\paragraph{Influence of Image Resolution}

\begin{table}[t]
\centering
\caption{Comparison of success rate between compressed and original images across different training stages and datasets.}
\label{tab:image_compression}
\begin{tabular}{lcccc}
\toprule
\multirow{2}{*}{Model} & \multicolumn{2}{c}{Compressed Images} & \multicolumn{2}{c}{Original Images} \\
\cmidrule(lr){2-3} \cmidrule(lr){4-5}
 & AndroidWorld & AndroidLab & AndroidWorld & AndroidLab \\
\midrule
Reasoning-Free SFT & 48.9 & 39.8 & 48.1 & 42.7 \\
Reasoning SFT      & 66.1 & 40.3 & 66.2 & 45.0 \\
\onlinerl (MobileRL-9B) & 75.8 & 46.8 & 80.2 & 53.6 \\
\bottomrule
\end{tabular}
\end{table}

In our preliminary experiments, we applied a uniform compression to images (maximum pixels per image = 500,000) across all three stages in order to accelerate training, yielding the initial results. The standard practice, however, is to use uncompressed images. For experiments based on Qwen2.5-VL-Instruct, we observed that this difference did not significantly affect performance while greatly accelerating training. In contrast, for experiments based on GLM-4.1V-9B-Base, we found notable accuracy differences between compressed and uncompressed settings, as shown in Table~\ref{tab:image_compression}. Therefore, we updated the reported results to those obtained with full-resolution images.

This change in accuracy is reasonable in the SFT stage, since the AndroidLab tasks contain more question-answering oriented tasks, where higher resolution images often make it easier to read screen content. In the RL stage, clearer images allow the agent to explore more thoroughly, leading to higher scores.

\section{XML Preprocessing for UI Representation}
\label{appendix:xml-processing}

The original XML from the Android accessibility service defines the layout and elements of the user interface, including all components on a page. As a result, it contains many nodes used solely for structural or layout purposes, which do not provide useful semantic information. Moreover, scrollable pages often contain more nodes than are visible on the screen, leading to the inclusion of many off-screen nodes.

\subsection{Removal of Off-Screen Nodes}
We first determine whether to retain off-screen nodes via the input parameter \texttt{remain\_nodes}:
\begin{itemize}
    \item \textbf{\texttt{remain\_nodes=True}}: Off-screen nodes are preserved, e.g., when summarizing the full page content without requiring scrolling.
    \item \textbf{\texttt{remain\_nodes=False}}: Off-screen nodes are removed to avoid interference during action simulation (e.g., tapping, scrolling).
\end{itemize}
In the original XML, a node is considered on-screen if its \texttt{bounds} property lies entirely within the screen dimensions $[0,0]$ to $[Window\_Height, Window\_Width]$ and is contained by its parent node. We check this condition recursively to identify on-screen nodes.

\subsection{Removal of Redundant Nodes}
Nodes that do not convey functional or semantic information are removed. A node is considered \emph{functional} if it satisfies at least one of the following:
\begin{itemize}
    \item Any of the boolean attributes is \texttt{True}: \texttt{checkable}, \texttt{checked}, \texttt{clickable}, \texttt{focusable}, \texttt{scrollable}, \texttt{long-clickable}, \texttt{password}, \texttt{selected}.
    \item The \texttt{text} or \texttt{content-desc} attribute is non-empty.
\end{itemize}
All nodes failing these criteria are considered redundant and are deleted.

\subsection{Attribute Simplification}
Attribute descriptions in the original XML are verbose and token-expensive. We simplify them as follows:
\begin{itemize}
    \item Keep only \texttt{True} values for the boolean functional attributes listed above (omit \texttt{False} values).
    \item Remove \texttt{index}, \texttt{resource-id}, and \texttt{package} (not useful for semantic understanding).
    \item For \texttt{class}, retain only the last component (e.g., \texttt{android.widget.FrameLayout} $\rightarrow$ \texttt{FrameLayout}).
    \item Merge \texttt{text} and \texttt{content-desc} attributes and display them separately.
    \item Retain \texttt{bounds} in full, as it indicates the node’s position on the page.
\end{itemize}

\subsection{Example Transformation}
Original node:
\begin{lstlisting}
<node index="0" text="Audio Recorder" 
      resource-id="com.dimowner.audiorecorder:id/txt_title" 
      class="android.widget.TextView" package="com.dimowner.audiorecorder" 
      content-desc="" checkable="false" checked="false" clickable="false" 
      enabled="true" focusable="false" focused="false" scrollable="false" 
      long-clickable="false" password="false" selected="false" 
      bounds="[221,1095][858,1222]" />
\end{lstlisting}

Simplified node:
\begin{lstlisting}
TextView;;Audio Recorder;[221,1095][858,1222]
\end{lstlisting}

\section{Data Collection}
\label{appendix:data-processing}

In this section, we present the composition and construction methodology of the training and testing data used at each stage, as well as their respective proportions.

\subsection{Data Distribution}

We present the source distribution of the first two stages of SFT data in Table~\ref{tab:sft-distribution}. 
In the RL stage, we utilized 2,000 tasks from AndroidWorld and 1,103 tasks from AndroidLab. Details of our data construction, deduplication, and reward acquisition methods are provided in a later section. We compared the performance of training solely on AndroidWorld with that of mixed training on both AndroidWorld and AndroidLab. The results varied across different model backbones: models based on Qwen2.5-VL-7B-Instruct performed better under the mixed-training setting, whereas models based on GLM-4.1V-9B-Base achieved superior results when trained only on AndroidWorld. Accordingly, we report the best results for each backbone.

\begin{table}[htbp]
    \centering
    \caption{Data distribution and labels across the two SFT training stages by steps.}
    \label{tab:sft-distribution}
    \begin{tabular}{lcccc}
        \toprule
        \textbf{Stage} & 
        \makecell[c]{\textbf{Android}\\\textbf{Control (Low)}} & 
        \makecell[c]{\textbf{Android}\\\textbf{Control (High)}} & 
        \makecell[c]{\textbf{Human Annotation}} & 
        \textbf{Total} \\
        \midrule
        Reasoning-Free SFT & 21.3k & 14.3k & 62.2k & 97.9k \\
        Reasoning SFT      & 7.2k  & 4.1k  & 12.2k & 23.6k \\
        \bottomrule
    \end{tabular}
    
\end{table}

\subsection{Human Annotation}

\label{appendix:human annotation}

Following the approach in~\cite{xu2024androidlabtrainingsystematicbenchmarking}, we collected a portion of human-annotation data and additionally employed model-driven self-exploration on the online interactive environment, performing online rollouts and selecting the correct action sequences for training. 

\paragraph{Privacy Protection}

In this work, the data collection process for model training inevitably involves capturing page screenshots and corresponding XML files. To address this, a privacy protection interface that does not retain raw data is applied before storage, ensuring that all sensitive information is filtered out in advance and that only sanitized data is preserved. Building upon this foundation, the overall scope of privacy protection is defined to cover both textual and visual information, with sensitive elements being systematically identified and anonymized. For textual data, the detection process primarily employs regular expression–based rules to recognize personal identifiers such as social media accounts, vehicle registration numbers, shipment tracking codes, order numbers, bank account details, email addresses, passport information, national identification numbers, and phone numbers. Once detected, these elements are replaced with standardized masking tokens that indicate the category of the information without revealing the actual content, thereby preserving semantic integrity while mitigating risks of exposure. For visual data, the privacy protection mechanism extends to the automatic detection of faces, QR codes, and text extracted from images through optical character recognition. Detected sensitive regions within images are then masked or obfuscated, producing new visual outputs in which private content is effectively concealed. Taken together, these measures form a comprehensive privacy-preservation strategy that integrates rule-based textual analysis with multimodal image processing, ensuring robust safeguards against the leakage of personal or identifying information throughout the data collection and annotation pipeline.

\subsection{AndroidLab}

\subsubsection{Data Processing}

The AndroidLab~\citep{xu2024androidlabtrainingsystematicbenchmarking} dataset consists of an online evaluation set containing 138 problems, all of which are equipped with verifiable evaluation mechanisms. In our study, we directly adopt the SR (Success Rate) metric provided by AndroidLab as the primary reward indicator. However, since AndroidLab does not include a corresponding verifiable training set, we construct one by means of a model-driven expansion strategy. Specifically, we utilize the applications included in AndroidLab to automatically generate candidate training tasks, which are subsequently subjected to manual verification to ensure both feasibility and non-overlap with the test set. Through this process, we compile a total of 1,103 training problems.

Since ground-truth test methods do not accompany these generated problems, it is not possible to rely on the same direct evaluation mechanism used for the test set. To address this issue, we train a reward model that serves as an auxiliary evaluator of correctness. This reward model is employed in two stages of our training pipeline: (i) during the Reasoning SFT stage, it functions as a filter to ensure the quality and validity of training data; and (ii) during the Reinforcement Learning stage, it provides a learnable reward signal that guides policy optimization.


\subsubsection{Reward Model for AndroidLab}
\label{appendix:reward-model}

Since the AndroidLab environment does not provide rule-based rewards for training data, we adopt a VLM-based reward model to supply reward signals during reinforcement learning. Specifically, we first execute all training and test tasks multiple times using medium-capability VLMs, including different versions of our models and proprietary VLMs, generating execution traces. Then, strong proprietary VLMs assign binary scores to each trace. For scoring, each model receives the full task screenshot sequence concatenated into a single image, along with step-by-step action descriptions, and is instructed to produce a reasoning process before outputting a score. For training-set tasks (non-rule-based), we use the majority vote among the three scores as the label and retain the reasoning; for test-set tasks, we use the existing rule-based reward as the label and keep only the model scores and reasoning consistent with it. This process yields several thousand labeled samples, which we use to fine-tune GLM-4.1V-9B-Thinking as the base model. After training, we evaluate the reward model on a curated set of 1000 AndroidLab traces with verified labels, selecting the best-performing version (86\% accuracy) for online reinforcement learning.

\paragraph{System Prompt for Trace Evaluation}
The following system prompt guides the VLM in determining whether an agent has successfully completed a task.

\begin{lstlisting}
You are an expert in determining whether a task 
has been successfully and completely completed. You will receive:

1. The task description.
2. Step-by-step page states in XML format.
3. The agent's action descriptions.
4. A single image containing screenshots of all steps.
   - Green rectangles and crosses mark tap regions and positions.
   - Green arrows indicate swipe directions.
   - Red text shows typed input.

Action formats and meanings:
- do(action="Tap", element=[x1,y1,x2,y2])
    Tap on the specified screen region; green cross is the tap point.
- do(action="Launch", app="xxx")
    Launch the specified app.
- do(action="Type", text="xxx")
    Enter the specified text (shown in red).
- do(action="Swipe", element=[x1,y1,x2,y2], direction="x", dist="x")
    Swipe in the indicated direction; green arrow shows the swipe path.
- do(action="Long_Pres", element=[x1,y1,x2,y2])
    Long press on the specified region.
- do(action="Back")
    Navigate back to the previous screen.
- finish(message="xxx")
    End the task with the given message.

Scoring rules:
If the task is fully and correctly completed,
output a score of 1; otherwise, 0.

Output format:
<analysis>
Step 1 analysis: <Your analysis>
Step 2 analysis: <Your analysis>
...
Final step analysis: <Your analysis>
</analysis>
<ans>
[Your score]
</ans>
\end{lstlisting}

\paragraph{Analysis}

\begin{figure*}[h!]
  \centering
  \includegraphics[width=0.8\linewidth]{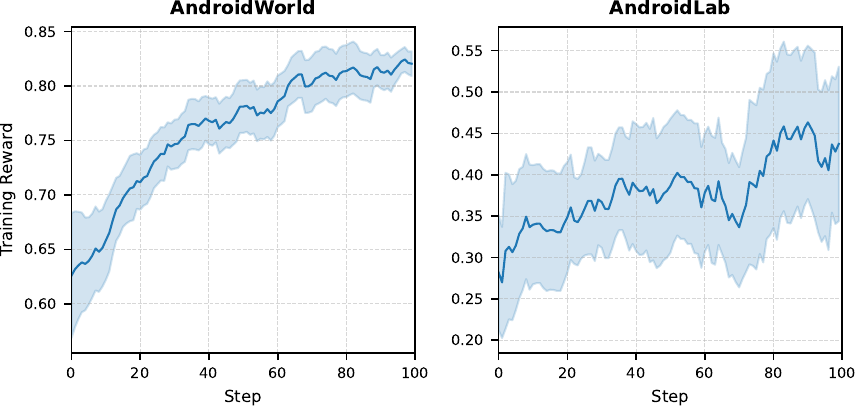}
  \vspace{-2mm}
  \caption{Trajectory-level rewards with 95\% CIs on training sets, with the reasoning SFT version of Qwen-2.5-VL-Instruct as backbone model. showing consistent performance growth. 
  }
  \label{fig:train_curve}
  \vspace{-3mm}
\end{figure*}

We compared the training curves of AndroidWorld and AndroidLab (trained simultaneously), as shown in Figure~\ref{fig:train_curve}. The training curve of AndroidWorld is smoother compared to that of AndroidLab, which we attribute to the use of a rule-based reward in AndroidWorld, whereas AndroidLab relies on a reward model. However, the reward model cannot fully match the accuracy of the rule-based reward. Therefore, we believe that evaluating whether a broader range of tasks has been successfully completed without relying on rule-based rewards remains an important challenge for future work.

\subsection{AndroidWorld}

\subsubsection{Data Processing}

The AndroidWorld dataset~\citep{rawles2024androidworlddynamicbenchmarkingenvironment} comprises 116 test problems, each equipped with verifiable reward signals. Since its native action space differs from ours, we adapt the operation code to align with our action space. AndroidWorld further provides randomized initialization parameters, which enable the generation of a large number of training tasks with verifiable rewards by replacing predefined substitutable components in the tasks (e.g., input content) and by varying the initial device states. To prevent overlap with the evaluation set, we remove all tasks whose specific content and initial states coincide with those in the test set. From the remaining pool, we curate 2,000 tasks to serve as the training set.

\subsection{AndroidControl}

\begin{table*}[t]
  \centering
  \caption{The correspondence between the AndroidControl action space and MobileRL action space, as well as the judgment details for each action.}
  \label{tab:AC_actions}
  \small
  \begin{tabularx}{\textwidth}{@{}l l X@{}} 
    \toprule
    \textbf{AC Action}   & \textbf{MobileRL Action}                                                         & \textbf{Evaluation Details}                                      \\
    \midrule
    click                & Tap                                                                         & Consider correct if the action type matches and the click point falls within the Tap bounding box.                                      \\
    long\_press          & Long Press                                                                  & Consider correct if the action type matches and the long\_press point falls within the Long Press bounding box.                         \\
    scroll               & Swipe                                                                       & Consider correct if the action type matches and the scroll direction corresponds to the Swipe direction.                       \\
    open\_app            & Launch                                                                      & Consider correct if the action type matches and the Launch app matches.                   \\
    input\_text          & Type                                                                        & Consider correct if the action type matches and the Type text F1 score with the input\_text text is higher than 0.5.                     \\
    navigate\_home       & Home                                                                        & Consider correct if the action type matches.                        \\
    navigate\_back       & Back                                                                        & Consider correct if the action type matches.                        \\    
    wait                 & Wait                                                                        & Consider correct if the action type matches.                         \\
    \bottomrule
  \end{tabularx}
\end{table*}

\subsubsection{Data Processing}


Each test case in the AndroidControl~\citep{li2024effects} dataset includes \texttt{episode\_id}, \texttt{goal}, \texttt{screenshots}, \texttt{accessibility\_trees}, \texttt{screenshot\_widths}, \texttt{actions}, \texttt{screenshot\_heights}, and \texttt{step\_instructions}. The format of input information and output actions differs from that of our action space, and we provide transformation details in Table~\ref{tab:AC_actions}. Certain transformations are required for an evaluation to be performed. The dataset consists of 8,444 test samples, 690 validation samples, and 74,714 training samples. 

\paragraph{Transformation from Accessibility Tree to XML}
We first transform the input information \texttt{accessibility\_tree} to XML. The XML obtained from mobile pages contains nodes with the following attributes: \texttt{index}, \texttt{resource-id}, \texttt{class}, \texttt{package}, \texttt{content-desc}, \texttt{checkable}, \texttt{checked}, \texttt{clickable}, \texttt{enabled}, \texttt{focusable}, \texttt{focused}, \texttt{scrollable}, \texttt{long-clickable}, \texttt{password}, \texttt{selected}, and \texttt{bounds}. We extract and set these corresponding attributes from each node of the accessibility tree. This ensures that the converted XML is consistent with the XML obtained directly from mobile pages. We recursively process through steps as follows:
\begin{itemize}
    \item Recursively process the three levels of \texttt{accessibility\_tree}, \texttt{window}, and \texttt{node}.
    \item Set the corresponding XML attribute for each attribute of the node.
    \item Compress the converted XML into compressed\_XML format.
\end{itemize}
Here is a transformation example:

Accessibility tree node:
\begin{lstlisting}
nodes {
      bounds_in_screen {
        right: 1080
        bottom: 2400
      }
      class_name: "android.widget.FrameLayout"
      package_name: "com.zoho.meeting"
      text_selection_start: -1
      text_selection_end: -1
      window_id: 1733
      is_enabled: true
      is_visible_to_user: true
      actions {
        id: 4
      }
      actions {
        id: 8
      }
      actions {
        id: 64
      }
      actions {
        id: 16908342
      }
    }
\end{lstlisting}

XML node:

\begin{lstlisting}
<node index="0" text="" resource-id="" class="android.widget.FrameLayout" package="com.zoho.meeting" content-desc="" checkable="false" checked="false" clickable="false" enabled="true" focusable="false" focused="false" scrollable="false" long-clickable="false" password="false" selected="false" bounds="[0,0][1080,2400]"/>
\end{lstlisting}

\paragraph{Transformation from Point to Bounding Box}
The grounding actions in AndroidControl use points. Since our grounding actions use bounding boxes rather than points, conversion of grounding coordinates is required. We attempt to match all UI boxes on a page. If the point falls within a UI box, the original click coordinates are converted to the coordinates of that UI box.

\paragraph{Evaluation Details}
Our evaluation metric follows~\cite{lian2025uiagileadvancingguiagents}, as it provides open-source complete evaluation code, as shown in Table~\ref{tab:AC_actions}. Notably, since our grounding actions use bounding boxes instead of points, we have modified the judgment logic for positioning operations. A grounding action is considered correct if the point coordinates of the original action fall within the UI box predicted by the model.
For the AndroidControl Low evaluation, we only conduct single-turn conversations with the agent. The provided information includes the instruction for the current step, the screenshot, and compressed XML data.
For the AndroidControl High evaluation, we maintain a history with the agent and replace incorrect actions. The provided information includes the total task, the screenshot for the current step, and compressed XML data.
\section{Limitation}

\paragraph{Reward Model} 
We train a reward model to gain reward signals during the training on the AndroidLab environment. The training data is obtained from strong proprietary VLMs through majority voting, but it is noteworthy that bias still exists in those data. Therefore, the reward model is inevitably born slightly biased from a perfect judge on agentic tasks, which brings about unsatisfying results during online RL in some cases. In particular, while \method-7B and all ablation studies are carried out with the existence of the reward model, the best model, \method-9B, is gained from online RL purely on the AndroidWorld environment. In observation of this, we conclude that a general and unbiased verification model is of significant importance and necessity, which is proposed as one aspect of our future work.

\paragraph{Difficulties with Public-Use Apps}
Our current work does not evaluate on out-of-domain public-use apps. Although the training data from AndroidControl and human annotations both include some tasks within public-use apps, we did not conduct testing in this setting. The main reason is the lack of a reproducible and efficient evaluation protocol in the open domain. Existing approaches typically require human execution of tasks and manual assessment of outcomes, which are costly and difficult to scale.  

Moreover, public-use apps often introduce practical challenges. Many require user login, cannot be run reliably on virtual machines, or lack mechanisms to restore application states, making online rollouts particularly challenging from an engineering perspective. We plan to explore solutions for training and evaluating on public-use apps in future work.

\paragraph{Limitation of prediction based on XML}
We observed that certain applications on the Android platform, especially those with large popularity and of comprehensive functions(e.g. TikTok, etc.), lack sufficient information about all interactable elements in their XML hierarchy data. To successfully perform complex tasks on these Apps, the current format we chose in model prediction, which is based on bounding box information contained in XML data, needs adaptation toward the general format built on solely visual information. This adaptation is, in the end, inevitable while challenging, so we leave it for our future work to obtain a generalized vision model with no need for XML data.

\paragraph{Training Time and Resource Consumption}

Another limitation of our work lies in the substantial resource demands of Android virtual devices. Each device requires more than 5 GB of memory and over 30 GB of storage, which restricts the number of rollouts that can be run concurrently. To address this, we employed four CPU machines with 1 TB of memory each, dedicated solely to rollouts. Although we implemented a distributed framework that supports multi-machine concurrent sampling, the system can only stably sustain up to 256 parallel rollouts, with each step taking more than 10 minutes.  

This constraint severely limits further scaling of training steps, as training 100 steps already takes more than 25 hours. In addition, large-scale image inference and network transmission introduce significant overhead, further increasing the computational burden.

\end{document}